\documentclass[accepted]{uai2025} % after acceptance, for a revised version; 

\usepackage{float}
\usepackage[american]{babel}
\usepackage{natbib}
\usepackage{mathtools}
\usepackage{booktabs}
\usepackage{amssymb}
\usepackage{amsthm}
\usepackage{bm}
\usepackage{bbm}
\usepackage{caption}
\usepackage{enumitem}
\usepackage{graphicx}
\usepackage{makecell}
\usepackage{microtype}
\usepackage{nicefrac}
\usepackage{subcaption}
\usepackage{url}
\usepackage{xcolor}
\usepackage{hyperref}
\usepackage[belowskip=0pt,aboveskip=6pt]{caption}
\newtheoremstyle{shortpropositionstyle}{\parskip}{0pt}{\it}{}{\bfseries}{.~}{0pt}{}
\theoremstyle{shortpropositionstyle}
\newtheorem{proposition}{Proposition}

\setlist{nosep,parsep=\parskip}
\setlist[itemize]{leftmargin=*}
\setlist[enumerate]{label=(\roman*),leftmargin=20pt}

\definecolor{deepblue}{HTML}{4C72B0}
\definecolor{deeporange}{HTML}{DD8452}
\definecolor{deepgreen}{HTML}{55A868}
\definecolor{deepred}{HTML}{C44E52}
\definecolor{deeppurple}{HTML}{8172B3}
\definecolor{deepbrown}{HTML}{937860}
\definecolor{deeppink}{HTML}{DA8BC3}
\definecolor{deepgray}{HTML}{8C8C8C}
\definecolor{deepyellow}{HTML}{CCB974}
\definecolor{deepcyan}{HTML}{64B5CD}
\definecolor{darkblue}{HTML}{001C7F}
\definecolor{darkorange}{HTML}{B1400D}
\definecolor{darkgreen}{HTML}{12711C}
\definecolor{darkred}{HTML}{8C0800}
\definecolor{darkpurple}{HTML}{591E71}
\definecolor{darkbrown}{HTML}{592F0D}
\definecolor{darkpink}{HTML}{A23582}
\definecolor{darkgray}{HTML}{3C3C3C}
\definecolor{darkyellow}{HTML}{B8850A}
\definecolor{darkcyan}{HTML}{006374}

\def\*#1{\bm{#1}}
\def\+#1{\mathrm{#1}}
\newcommand{\justify}[1]{{#1\parfillskip=0pt\par}}

\title{Transparent Trade-offs between Properties of Explanations}

\author[1]{Hiwot Belay Tadesse}
\author[1]{Alihan H\"uy\"uk}
\author[2]{Yaniv Yacoby}
\author[1]{Weiwei Pan}
\author[1]{Finale Doshi-Velez}
\affil[1]{John A. Paulson School of Engineering and Applied Sciences, Harvard University, Cambridge, MA, USA}
\affil[2]{Wellesley College, Wellesley, MA, USA}
\begin{document}
\maketitle

\begin{abstract}
    When explaining machine learning models, it is important for explanations to have certain properties like faithfulness, robustness, smoothness, low complexity, etc. However, many  properties are in tension with each other, making it challenging to achieve them simultaneously. For example, reducing the complexity of an explanation can make it less expressive, compromising its faithfulness. The ideal balance of trade-offs between properties tends to vary across different tasks and users. Motivated by these varying needs, we aim to find explanations that make \textit{optimal trade-offs} while allowing for \textit{transparent control} over the balance between different properties. Unlike existing methods that encourage desirable properties implicitly through their design, our approach optimizes explanations explicitly for a linear mixture of multiple properties. By adjusting the mixture weights, users can control the balance between those properties and create explanations with precisely what is needed for their particular task.
\end{abstract}

%%%%%%%%%%%%%%%%%%%%%%%%%%%%%%%%%%%%%%%
%%%%%%%%%%%%%%%%%%%%%%%%%%%%%%%%%%%%%%%
%%%%%%%%%%%%%%%%%%%%%%%%%%%%%%%%%%%%%%%
%%%%%%%%%% INTRODUCTION
%%%%%%%%%%%%%%%%%%%%%%%%%%%%%%%%%%%%%%%
%%%%%%%%%%%%%%%%%%%%%%%%%%%%%%%%%%%%%%%
%%%%%%%%%%%%%%%%%%%%%%%%%%%%%%%%%%%%%%%

\section{Introduction}

\looseness-1
When explaining machine learning models, it is desirable for our explanations to satisfy various \textit{properties}. For instance, explanations should be \textit{faithful} and accurately reflect the actual computations carried out by the underlying model \citep{yeh2019fidelity}. We might also want them to be \textit{robust} so that similar inputs result in similar explanations \citep{alvarez2018towards}, or \textit{smooth} so that similar dimensions in the input are assigned similar values in the explanation \citep{ajalloeian2022smoothed}. Some applications may require explanations to be simple with low \textit{complexity} \citep{ijcai2020p417}. Accordingly, a range of metrics has been proposed to evaluate these
%different
properties \citep{chen2022makes,wang2024multi}.

While all of these properties can be useful, many of them are in tension with each other, making it difficult to achieve them simultaneously. For instance, reducing the complexity of an explanation often makes it less expressive, which can undermine its faithfulness \citep{ijcai2020p417}. Previous work has shown similar trade-offs between faithfulness vs.\ robustness \citep{tan2023robust}, faithfulness vs.\ sensitivity \citep{bansal2020sam}, and faithfulness vs.\ homogeneity \citep{balagopalan2022road}.

\begin{table}
    \centering
    \caption{Different tasks require explanations with different properties. For instance, when auditing models, explanations need to be faithful no matter how complex so that users can investigate every detail and catch the smallest breaches in regulation. Meanwhile, counterfactual reasoning involves extrapolating a model's behavior to new cases, where robust explanations that capture global trends might be more useful than explanations faithful to local variations.}%
    \label{tab:intro}%
    \small%
    \resizebox{\linewidth}{!}{
        \begin{tabular}{@{}ll@{~}c@{~}l@{~}l@{}}
             \toprule
             \bf Task & \multicolumn{4}{l}{\bf Property Preference} \\
             \midrule
             Auditing models & Faithfulness & $\succ$ & Complexity & \citep{nofshin2024sim2real} \\
             Counterfactual reasoning & Robustness & $\succ$ & Faithfulness & \citep{nofshin2024sim2real} \\
             Detecting model biases & Smoothness & $\succ$ & Faithfulness & \citep{colin2022cannot} \\
             \bottomrule
        \end{tabular}}%
\end{table}

Not only these trade-offs exist, but it is also the case that different tasks or different users require different balances among these properties \citep{zhou2021evaluating,liao2022connecting,nofshin2024sim2real}. For instance, users who work more closely with AI might prefer more faithful explanations despite their complexity, while non-AI experts might prefer explanations that are shorter and clearer instead \cite{wang2024multi}. Likewise, studies show that users perform better with faithful explanations when auditing black-box models for compliance \citep{nofshin2024sim2real}, robust explanations when performing counterfactual (``what-if'') reasoning \citep{nofshin2024sim2real}, or smooth explanations when trying to detect model biases (see Table~\ref{tab:intro}) \citep{colin2022cannot}.

Motivated by these varying user needs, we aim to provide a method for finding explanations that make \textit{optimal} trade-offs between different properties, where the balance between each property can be controlled \textit{transparently}. We focus specifically on feature attribution explanations \citep[e.g.][]{ribeiro2016should,lundberg2017unified, smilkov2017smoothgrad,selvaraju2020grad}. This popular type of explanation assigns a contribution score to each input feature, capturing the impact of that feature on the final output. We achieve our goal by directly optimizing these contribution scores for a linear mixture of properties, where the mixture weights can be adjusted freely by the user.

We derive computationally efficient ways to perform such optimization in both inductive and transductive settings. In the inductive setting, we show that optimizing a combination of faithfulness, robustness, and smoothness can be expressed as an equivalent Gaussian process (GP) inference problem. While the inductive setting is more generally applicable, the transductive setting allows us to consider an even wider range of properties and efficient optimization strategies. In particular, in the transductive setting, we consider optimizing for complexity and alternative formulations of faithfulness and robustness from the literature.

\begin{figure}
    \centering
    \includegraphics[width=.95\linewidth]{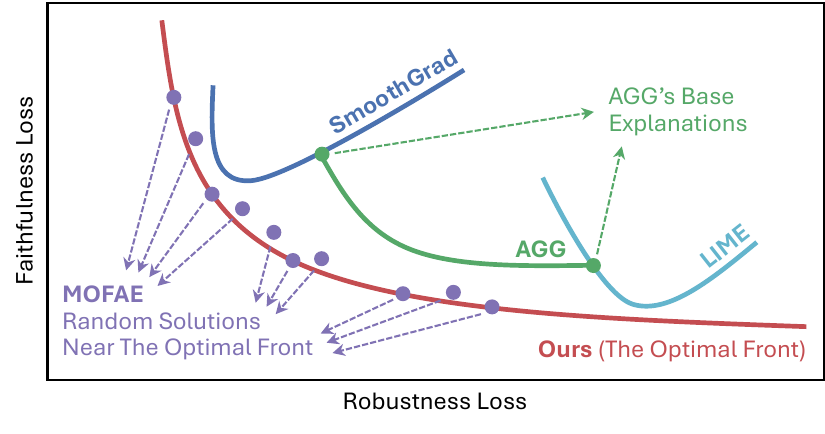}%
    \caption{
        \textbf{Summary of Related Work.}
        Consider the trade-off between two properties. Methods without explicit optimization (like \textcolor{darkblue}{\bf SmoothGrad} and \textcolor{darkcyan}{\bf LIME}) fail to reliably produce optimal explanations. Varying their hyperparameters leads to different explanations, but does not consistently balance one property againts the other. \textcolor{darkgreen}{\bf AGG} can offer more direct control over the trade-off by aggregating multiple explanations, but its span and optimality are limited by the quality of its base explanations. \textcolor{darkpurple}{\bf MOFAE} can generate a random set of near-optimal explanations, but provides little control over where exactly these explanations land along the optimality front (making it difficult to fine-tune the balance for an individual explanation). In contrast, \textcolor{darkred}{\bf our approach} finds optimal explanations with transparent control across the entire optimality front.}%
    \label{fig:related}%
\end{figure}

\textbf{Related Work: Limitations of Current Methods.}
Most feature attribution methods tend to target a specific property when generating explanations. For instance, SmoothGrad \citep{smilkov2017smoothgrad} improves robustness by averaging gradients around input points. Meanwhile, LIME \citep{ribeiro2016should} prioritizes faithfulness by fitting local linear models at individual input points. Like SmoothGrad and LIME, methods are typically designed as forward computations, where the desired property is encouraged heuristically through the design of those computations. Explanations are almost never optimized directly for a target property or for a desired trade-off between multiple properties.

As a consequence, such methods often make implicit trade-offs against properties other than the one targeted. For instance, methods like LIME or SHAP \citep{lundberg2017unified}, which focus on faithfulness, have been shown to generate explanations that are not always robust \citep{alvarez2018towards,ghorbani2019interpretation,slack2020fooling}. Similarly, methods like SmoothGrad and GradCAM \citep{selvaraju2020grad}, which focus on robustness, can yield explanations that lack faithfulness \citep{adebayo2018sanity}.

These implicit trade-offs make it impossible to control the balance between competing properties in a transparent manner. To give an example, \citet{tan2023robust} demonstrated for SmoothGrad that the balance between faithfulness and robustness is sensitive to a hyperparameter, $\delta$, but how exactly $\delta$ affects this balance remains unclear (as we will see in the experiments, varying $\delta$ can lead to unintuitive changes in faithfulness and robustness). Furthermore, without explicit optimization, the resulting trade-off may not even be optimal or the method might fail to produce the intended property altogether. Frameworks for recommending explanations tailored to specific user needs \citep[e.g.][]{cugny2022autoxai} often rely on tuning the hyperparameters of existing explanation methods, inheriting the same limitations.

In order to produce a desired property more reliably, \citet{ijcai2020p417,decker2024provably} proposed aggregating explanations generated by multiple methods. While the properties induced by individual methods can be unpredictable, with sufficient diversity, finding a desirable aggregation can be possible. The algorithm of \citet{decker2024provably}, AGG, searches for the best convex combination of some base explanations to optimize either for faithfulness or for robustness. AGG can be extended to target any arbitrary trade-off between these two properties as well, offering transparent control, however, the span and optimality of achievable trade-offs would remain constrained by the properties of the base explanations used (see Figure~\ref{fig:related}).

\citet{wang2024multi} proposed a framework where explanations are directly optimized for multiple properties simultaneously (rather than a single mixture as in our approach). Their method, MOFAE, uses a genetic algorithm to generate a collection of explanations, each achieving a different but optimal trade-off. However, since this collection is generated through a stochastic process, MOFAE is ineffective at targeting a particular balance of properties or at fine-tuning the balance achieved by an individual solution (see Figure~\ref{fig:related}).

\begin{table*}
    \centering
    \caption{
        \textbf{Summary of Property Definitions.} We consider faithfulness (gradient matching), robustness (average differences), and smoothness for both inductive and transductive settings. While the inductive setting is more generally applicable, the transductive setting allows us to consider additional definitions of faithfulness (function matching) and robustness (maximum difference) as well as complexity. $^{\dagger}$$\smash{s(\*x,\*x')}$ and $\smash{S_{nn'}\doteq s(\*x_n,\*x_{n'})}$ quantify the similarity between two inputs. $^{\ddagger}$$\smash{\tilde{S}_{dd'}}$ quantifies the similarity between two input dimensions.}
    \label{tab:properties}
    \newcommand{\cellmathstyle}{\textstyle}
    \resizebox{\linewidth}{!}{
        \begin{tabular}{@{}lc@{~~}lc@{~~}l@{}}
            \toprule
            \bf Property & \multicolumn{2}{l}{\bf Inductive Definition} & \multicolumn{2}{l}{\bf Transductive Definition} \\
            \midrule
            Faithfulness \textit{(Gradient Matching)}
                & \textbf{(a)} & $\cellmathstyle \mathcal{L}_{\texttt{F-grad}}(E)=\int_{\Omega}\|E(\*x)-\nabla f(\* x)\|_2^2 d\*x$
                & \textbf{(d)} & $\cellmathstyle \mathcal{L}_{\texttt{F-grad}}(W_E)=\sum\nolimits_n\|\*w_{E_n}-\nabla f(\*x_n)\|_2^2$ \\
            Faithfulness \textit{(Function Matching)}
                & -- & & \textbf{(e)} & $\cellmathstyle \mathcal{L}_{\texttt{F-func}}(W_E)=\sum\nolimits_n|(\*w_{E_n})^{\top}\*x_n-f(\*x_n)|^2$ \\
            % \midrule
            Robustness \textit{(Average Difference)} $^{\dagger}$
                & \textbf{(b)} & $\cellmathstyle \mathcal{L}_{\texttt{R-avg}}(E)=\iint_{\Omega}\|\+E(\*x)-\+E(\*x')\|_2^2\; s(\*x,\*x')\; d\*xd\*x'$
                & \textbf{(f)} & $\cellmathstyle \mathcal{L}_{\texttt{R-avg}}(W_E)=\sum\nolimits_{n,n'}\|\bm{w}_{E_n}-\bm{w}_{E_{n'}}\|_2^2\; S_{nn'}$ \\
            Robustness \textit{(Maximum Difference)}
                & -- & & \textbf{(g)} & $\cellmathstyle \mathcal{L}_{\texttt{R-max}}(W_E)=\sum\nolimits_n\max\nolimits_{n'}\|\*w_{E_n}-\*w_{E_{n'}}\|_2^2$ \\
            % \midrule
            Smoothness $^{\ddagger}$
                & \textbf{(c)} & $\cellmathstyle \mathcal{L}_{\texttt{S}}(E)=\int_{\Omega}\sum\nolimits_{d,d'}|E_d(\*x)-E_{d'}(\*x)|^2\, \tilde{S}_{dd'}\, d\*x$
                & \textbf{(h)} & $\cellmathstyle \mathcal{L}_{\texttt{S}}(W_E)=\sum\nolimits_{n,d,d'}|(\*w_{E_n})_d-(\*w_{E_n})_{d'}|^2\, \tilde{S}_{dd'}$ \\
            % \midrule
            Complexity
                & -- & & \textbf{(i)} & $\cellmathstyle \mathcal{L}_{\texttt{C}}(W_E)=\sum\nolimits_n\|\*w_{E_n}\|_1$ \\
            \bottomrule
        \end{tabular}}
\end{table*}

\textbf{Contributions.} We make three contributions: (i)~ We propose a new method called \textbf{POE (Property Optimized Explanations)}, which directly optimizes for a combination of desired properties. POE allow for fine-grained control over the trade-offs between competing objectives, enabling users to create explanations with precisely the balance of properties needed for their task. (ii)~We demonstrate that popular feature attributions methods (e.g.\ SmoothGrad, LIME) not only fail to reliably generate explanations with optimal properties, but also lack mechanisms for controlling the prioritization of properties. Among recent frameworks, AGG brings some control but does not ensure optimality, and MOFAE finds optimal explanations but does not allow for fine-grained control. (iii)~Finally, we demonstrate through a variety of experiments that POE consistently yields explanations with optimal properties and easy-to-adjust trade-offs.

%%%%%%%%%%%%%%%%%%%%%%%%%%%%%%%%%%%%%%%
%%%%%%%%%%%%%%%%%%%%%%%%%%%%%%%%%%%%%%%
%%%%%%%%%%%%%%%%%%%%%%%%%%%%%%%%%%%%%%%
%%%%%%%% PROBLEM FORMULATION
%%%%%%%%%%%%%%%%%%%%%%%%%%%%%%%%%%%%%%%
%%%%%%%%%%%%%%%%%%%%%%%%%%%%%%%%%%%%%%%
%%%%%%%%%%%%%%%%%%%%%%%%%%%%%%%%%%%%%%%

\section{Problem Formulation}
\label{sec:formulation}

\textbf{Setting.}
We consider the problem of explaining some fixed function $f:\mathbb{R}^D\to\mathbb{R}$. We assume that we can query the function for an output $y = f(\*x)$ for any input $\*x$ as well as a gradient $\nabla f(\*x)$. Depending on how explanations are requested, we consider two different settings: (i) the \textit{inductive setting}, where the requests for explanations arrive in an online fashion (i.e.\ one input point $\*x$ at a time), and (ii) the \textit{transductive setting}, where we are given the full set of input points $\{\*x_n\}_{n=1}^N$ that we will need to explain in advance. While the inductive setting is more generally applicable, it is also a more challenging setting (because properties like robustness that rely on simultaneously optimizing explanations at multiple inputs requires us to reason about the entire input domain); the assumption of the transductive setting will allow us to optimize for a wider range of properties. % than the inductive setting.

\textbf{Explanations.}
We restrict our focus to local explanations that are based on feature-attributions. 
% FDV: Okay, but can we say sooner? Also, feature attributions are very common/popular/somethng?
This is so that we can give concrete definitions of properties such as faithfulness, robustness, and smoothness as an exemplar of our approach. In local feature attribution, an explanation $\*E\in\mathbb{R}^D$ is a vector assigning an attribution score to each component of some input point $\*x$. In the inductive settings, these explanations are given by a function $E:\mathbb{R}^D\to\mathbb{R}^D$ (from input to explanation). In the transductive setting, we do not assume an explicit explanation function, rather, for each input $\*x_n$, we denote the explanation at $\*x_n$ by $\*w_{E_n}\in\mathbb{R}^D$ and the matrix of explanations for the $N$ inputs as $W_E\in\mathbb{R}^{N\times D}$.

\textbf{The General Problem.}
We are given a set of desired properties $\{\text{prop}_i\}$, each characterized by a loss function $\mathcal{L}_{\text{prop}_i}$. In the inductive setting, these losses are functions of $E$, and in the transductive setting, they are functions of $W_E$. Our objective is to find the explanation function $E$---or the explanation matrix $W_E$---that optimizes for the given properties (i.e.\ minimizes their characteristic loss functions):
\begin{align}
    \textstyle E^* = \+{argmin}_E\sum\nolimits_i \lambda_{\text{prop}_i}\mathcal{L}_{\text{prop}_i}(E;f) \label{eqn:big_problem}
\end{align}
where $\lambda_{\text{prop}_i}$ are weights that provide a transparent, intuitive way to manage trade-offs between different properties. % in the optimization.

% FDV: Is it all \textbf below for space? (the specific properties hierarchically belong under this header, right?) Maybe there's a way to still convey that?
\section{Property Definitions}

In this paper, we optimize for a range of common properties that have been shown to be useful for downstream tasks in literature. In existing literature, the same conceptual property is mathematically formalized in many different ways \citep[][e.g.\ \citet{alvarez2018towards} vs.\ \citet{yeh2019fidelity} for robustness]{chen2022makes}. We focus on the properties of faithfulness, robustness, smoothness, and complexity.
We choose to work with these four properties because they are in tension, that is, it is not possible to maximize all four unless the function has constant gradients (i.e.\ it is linear). Thus, it is necessary for any explanation method to manage the trade-off between faithfulness, robustness, smoothness, and complexity in ways that are appropriate for specific tasks.
We consider formalizations of these properties that have been well studied in literature as well as lend themselves to efficient optimization. Table~\ref{tab:properties} summarizes all the properties we consider and their corresponding losses.

\textbf{Faithfulness.}
Faithfulness quantifies the extent to which an explanation accurately reflects the behavior of the function that is being explained. The class of explanations wherein one computes the marginal contribution of each dimension, by using small perturbations or by analytically computing input gradients, can be seen as optimizing faithfulness formalized as gradient matching \citep[Table~\ref{tab:properties}a,][]{baehrens2010explain, simonyan2013deep}:
\begin{align}
    \textstyle \mathcal{L}_{\texttt{F-grad}}(E) = \int_{\Omega} \|E(\*x)-\nabla f(\* x)\|_2^2 d\*x
    \label{eq:gradient_matching_faithfulness}
\end{align}
Many faithfulness metrics proposed in the literature can be reduced to gradient matching under certain settings. For example, the faithfulness metric in \citet{tan2023robust} is equivalent to gradient matching, when the perturbation is Gaussian and the similarity measure is Euclidean distance.
A number of metrics define faithfulness as matching the function with linear approximations based on the explanation, e.g.\ local-fidelity \citep{yeh2019fidelity}, local-accuracy \citep{tan2023robust}, loss-based-fidelity \citep{balagopalan2022road}, see $\mathcal{L}_{\texttt{F-func}}$ in Table~\ref{tab:properties}e. We will consider this alternative formalization in the transductive setting.

\textbf{Robustness.}
Robustness quantifies the extent to which an explanation varies with respect to the input. A common way to formalize robustness as a metric is to take the weighted sum of pairwise differences between explanations for different inputs in the data. More formally, in the inductive setting, this can be written (Table~\ref{tab:properties}b):
\begin{align}
    \textstyle \mathcal{L}_{\texttt{R-avg}}(E)= \iint_{\Omega} \|\+E(\*x) - \+E(\*x')\|_2^2\; s(\*x,\*x')\; d\*xd\*x'
    \label{eq:robustness}
\end{align}
where $s(\*x,\*x')\in\mathbb{R}$ is a weight that quantifies the similarity between $\*x$ and $\*x'$.
Metrics like (Lipschitz) local-stability \citep{alvarez2018towards, wang2020smoothed} are instances of the above equation, where we set the weights $s(\*x,\*x')$ to be a function of $\|\*x-\*x'\|_2^2$. Notably, our robustness loss depends on the similarity of the explanation at input $\*x$ to the explanation at input $\*x'$.  Thus, optimizing the robustness of an explanation for even \emph{one} input requires assigning explanations to \emph{many} other inputs.  By explicitly optimizing explanations, our approach will ensure that all explanation assignments are consistent, a property not present in most existing explanation methods. We detail our choice of $s$ in the appendix.
Rather than averaged differences, metrics like max-sen\-si\-tiv\-i\-ty \citep{yeh2019fidelity} capture a notion of robustness defined by the maximum difference between the explanations at ``neighboring'' input points, see $\mathcal{L}_{\texttt{R-max}}$ in Table~\ref{tab:properties}g. We will consider this alternative formalization in the transductive setting.

\textbf{Smoothness.}
Smoothness or ``internal robustness" captures the variability across different dimensions of an explanation that is assigned to a fixed input. Similar to robustness, suppose we are given a similarity matrix $\tilde{S}_{dd'}$ that tells how similar an input dimension $d\in[D]$ is to another input dimension $d'\in[D]$ (for instance, when inputs are images, this similarity can be related to the distance between two pixels). Then, we define the smoothness loss in a form similar to robustness (Table~\ref{tab:properties}c):
\begin{align}
    \textstyle \mathcal{L}_{\texttt{S}}(E) = \int_{\Omega}\sum\nolimits_{d,d'} |E_d(\*x)-E_{d'}(\*x)|^2\, \tilde{S}_{dd'}\, d\*x
\end{align}

\textbf{Complexity.}
A common way to formalize explanation complexity is via sparsity, i.e.\ by counting the number of non-zero weights in the explanations. In the transductive setting, this can be captured through the $\ell_1$-norm of explanations (Table~\ref{tab:properties}i): $\mathcal{L}_{\texttt{C}}(W_E) = \sum_n \|\*w_{E_n}\|_1$, which we prefer over $\ell_0$-norm for efficient optimization.

%%%%%%%%%%%%%%%%%%%%%%%%%%%%%%%%%%%%%%%
%%%%%%%%%%%%%%%%%%%%%%%%%%%%%%%%%%%%%%%
%%%%%%%%%%%%%%%%%%%%%%%%%%%%%%%%%%%%%%%
%%%%% INDUCTION: OPTIMIZING EXPLANATION FUNCTIONS
%%%%%%%%%%%%%%%%%%%%%%%%%%%%%%%%%%%%%%%
%%%%%%%%%%%%%%%%%%%%%%%%%%%%%%%%%%%%%%%
%%%%%%%%%%%%%%%%%%%%%%%%%%%%%%%%%%%%%%%

\vspace{-6pt}
\section{Induction: Optimizing Explanation Functions}
\label{sec:inductive}

\justify{
    In this section, we tackle the inductive version of our problem: learning explanation functions that are optimized for a desired mixture of explanation properties. We instantiate the general problem in Equation~\ref{eqn:big_problem} with three properties: faithfulness (\texttt{F-grad}), robustness (\texttt{R-avg}) and smoothness (\texttt{S}). Then, we drive a computationally efficient and mathematically consistent solution based on GP inference.}

\textbf{Instantiation.}
When linearly combined, the losses corresponding to these three properties ($\mathcal{L}_{\texttt{F-grad}}$, $\mathcal{L}_{\texttt{R-avg}}$, $\mathcal{L}_{\texttt{S}}$ in Table~\ref{tab:properties}a--c) lead to the following optimization problem:
\begin{align}
    E^* &= \+{argmin}_E\, \textstyle\int_{\Omega} \lambda_{\texttt{F-avg}} \|E(\*x) - \nabla f(\*x) \|^2_2 d\*x \nonumber \\
    &\hspace{6pt} + \lambda_{\texttt{R-avg}} \textstyle\iint_{\Omega} \|E(\*x) - E(\*x')\|^2_2 s(\*x,\*x') d\*xd\*x' \label{eqn:obj_ind_ex} \\
    &\hspace{6pt} + \lambda_{\texttt{S}} \textstyle\int_{\Omega} \sum\nolimits_{d,d'} |E_d(\*x)-E_{d'}(\*x)|^2 \tilde{S}_{dd'} d\*x \nonumber
\end{align}
Here, the robustness loss in Equation~\ref{eqn:obj_ind_ex} is what makes optimization challenging: When assigning an explanation even to one input, we must ensure that the explanation is similar to explanations for nearby inputs; those explanations, to be optimal, must in turn also be faithful, robust, and smooth. To get a consistent solution, we must reason over the entire space of explanation functions $E(\*x)$. This challenge is what sets the inductive setting apart from the transductive setting, where we only need to optimize a finite set of explanations.

\textbf{Equivalence to Gaussian Process Regression.}
We overcome the above challenge via Gaussian process (GP) regression, which provides a theoretically grounded and computationally tractable framework for inference in function spaces. Below, we show that the solution to the optimization problem in Equation~\ref{eqn:obj_ind_ex} can be expressed as the maximum aposteriori function of a multi-output GP, where gradients~$\nabla f(\*x)$ can be viewed as observations of some latent explanation function~$E(\*x)$, and the similarity measures $s,\tilde{S}$ can be written as the inverse of a multi-output kernel. Viewed as GP regression, we can drive a consistent, analytic solution to Equation~\ref{eqn:obj_ind_ex} for any number of query points.

Formally, let us start with the following GP prior over explanation functions $E(\*x)$:
\begin{align}
    E \sim \+{GP}(m,K),\quad m(\*x)=\mu\*1_D,\quad \mu\sim\mathcal{N}(0,\sigma^2) \label{eqn:gp}
\end{align}
where the mean $m$ has a constant value $\mu\in\mathbb{R}$ for all inputs and all dimensions, and the kernel $K$ is a matrix-valued function such that $\+{cov}(E_d(\*x),E_{d'}(\*x'))=K_{dd'}(\*x,\*x')$ with its inverse denoted as $K^{-1}$. We interpret the gradient $\nabla f$ as a random observation of some latent $E$, which determines the likelihood for posterior inference:
\begin{align}
    \nabla f(\*x) \sim \mathcal{N}(E(\*x),\varsigma^2 I) \label{eqn:gp_likelihood}
\end{align}
Then, we have the following equivalence:

\begin{proposition}
    \label{prop}
    For $\{\*x_n\}_{n=1}^N$, $\*x_n\sim\text{Uniform}(\Omega)$, and $\varsigma^2=1/N$, the maximum aposteriori function
    \begin{align}
        \+{argmax}_{E}~ p(E|\{\*x_n,\nabla f(\*x_n)\}) \label{eqn:prop-gp}
    \end{align}
    approaches to the optimal explanation function
    \begin{align}
        \begin{split}
            \+{argmin}_{E}~ \textstyle \int_{\Omega}\|&E(\*x)-\nabla f(\*x)\|_2^2 d\*x \\
            - \textstyle\iint_{\Omega}\sum_{d,d'}|&E_d(\*x)-E_{d'}(\*x')|^2K^{-1}_{dd'}(\*x,\*x')d\*xd\*x' \label{eqn:prop-objective}
        \end{split}
    \end{align}
    as $\sigma^2\to\infty$ and $N\to\infty$.
\end{proposition}

\textit{Proof.} The complete proof can be found in the appendix. However, the key insight is to write the inductive optimization in Eq.~\ref{eqn:prop-objective} as the limiting case of a transductive optimization, which allows us to manipulate it algebraically to show its equivalence to the inference problem in Eq.~\ref{eqn:prop-gp}. \qed

When the similarity measures in Equation~\ref{eqn:obj_ind_ex} are precision function corresponding to kernels, that is $s(\*x,\*x')=-k^{-1}(\*x,\*x')$ and $\tilde{S}=-\tilde{K}^{-1}$ for some $k,\tilde{K}$, we can define the GP kernel $K$ in Equation~\ref{eqn:gp} such that
\begin{align}
    \begin{split}
        K_{dd'}^{-1}(\*x,\*x') &= \textstyle \smash{\frac{\lambda_{\texttt{R-avg}}}{\lambda_{\texttt{F-grad}}}} \cdot k^{-1}(\*x,\*x') \cdot \mathbbm{1}\{d=d'\} \\
        &\hspace{6pt} + \textstyle \smash{\frac{\lambda_{\texttt{S}}}{\lambda_{\texttt{F-grad}}}} \cdot \mathbbm{1}\{\*x=\*x'\} \cdot \tilde{K}^{-1}_{d}
    \end{split} \label{eqn:gp_kernel}
\end{align}
Then, the objective in the proposition (Equation \ref{eqn:prop-objective}) becomes equivalent to the objective in our explanation optimization problem (Equation~\ref{eqn:obj_ind_ex}). Thus, the solution to our optimization problem can be computed via GP inference (via Proposition~\ref{prop}). In particular, we can find efficient solutions to our optimization problem through approximate GP inference by using a set of \textit{inducing points} $\{\*x_n\}$. In our experiments, we investigate the sensitivity of our approach to selection of inducing points (specifically to their size $N$, and additionally in the appendix, to their distribution).

%%%%%%%%%%%%%%%%%%%%%%%%%%%%%%%%%%%%%%%
%%%%%%%%%%%%%%%%%%%%%%%%%%%%%%%%%%%%%%%
%%%%%%%%%%%%%%%%%%%%%%%%%%%%%%%%%%%%%%%
%%%% TRANSDUCTION: OPTIMIZING EXPLANATIONS ON FIXED SETS
%%%%%%%%%%%%%%%%%%%%%%%%%%%%%%%%%%%%%%%
%%%%%%%%%%%%%%%%%%%%%%%%%%%%%%%%%%%%%%%
%%%%%%%%%%%%%%%%%%%%%%%%%%%%%%%%%%%%%%%

\section{Transduction: Optimizing Explanations for Fixed Inputs}
\label{sec:trans_opt}

Optimization in the inductive setting is generally challenging because the explanation function must be consistent across all possible inputs. In this section, we describe how optimization can be made even more efficient in the transductive setting, where we have in advance, a fixed set of inputs $\*x_n$ that we want to explain. In particular, we identify specific classes of properties for which the optimization problem in the transductive setting can be solved analytically or efficiently, and with guarantees, by recasting these problems as linear or quadratic programs.

% \textbf{Connecting the Inductive Setting with Quadratic Programming.} 
\textbf{Connection to the Inductive Setting.}
When we instantiate Equation~\ref{eqn:big_problem} for the transductive setting using the same definitions of faithfulness (\texttt{F-grad}), robustness (\texttt{R-avg}), and smoothness (\texttt{S}) from Section \ref{sec:inductive}, the optimization problem defines a quadratic program: 
\begin{align}
     W_E^* &= \textstyle \+{argmin}_{W_E}\, \lambda_{\texttt{F-grad}}\sum\nolimits_n \|\*w_{E_n} - \nabla f(\mathbf{x}_n)\|^2_2 \nonumber \\[-1pt]
    &\hspace{6pt} + \textstyle \lambda_{\texttt{R-avg}}\; \sum\nolimits_{n,n'} \|\*w_{E_n} - \*w_{E_{n'}} \|^2_2S_{nn'} \label{eqn:obj_trans_1} \\[-1pt]
    &\hspace{6pt} + \textstyle \lambda_{\texttt{S}}\; \sum\nolimits_{n,d,d'} |(\*w_{E_n})_d-(\*w_{E_n})_{d'}|^2\tilde{S}_{dd'} \nonumber
\end{align}
Furthermore, we note that when we choose the similarity measures $S$ and $\tilde{S}$ to be the negative precision matrices for some kernels $K$ and $\tilde{K}$, we can still take advantage of Proposition~\ref{prop} (when $N$ is large!) and use GP inference to find a solution. Doing so, the faithfulness objective follows from the Gaussian likelihood in Equation~\ref{eqn:gp_likelihood}, and the robustness and smoothness objectives are determined by our choice of kernel in the prior (Equation~\ref{eqn:gp}). This connection to GP inference provides theoretical grounding for the multi-property optimization problem in Equation~\ref{eqn:obj_trans_1}: Even though we are optimizing these properties for a fixed set of inputs, the resulting explanations are consistent with the wider, inductive notions of faithfulness, robustness, and smoothness.

%Specifically, the objective function in Equation~\ref{eqn:obj_trans_1} naturally results when we estimate the function $\nabla f$ from a given dataset using Bayesian regression and a GP prior -- the faithfulness objective follows from the Gaussian likelihood and the robustness objective is determined by our choice of kernel in the prior. Connection to Bayesian regression provides theoretical grounding for the multi-property optimization problem in Equation~\ref{eqn:obj_trans_1}. 

% FDV: Overall, wonder if we can make this section more salient to emphasize the wider range we solve for and requiring more than linear systems
\textbf{Additional Properties.}
In the transductive setting, we have the ability to efficiently optimize a large number of additional formalizations of properties by recasting them as linear or quadratic programs (as we no longer need the GP framework to keep track of functions). For example, we consider one additional formalization of faithfulness (\textit{function matching}, \texttt{F-func} in Table~\ref{tab:properties}e), one additional formalization of robustness (\textit{maximum difference}, \texttt{R-max} in Table~\ref{tab:properties}g) and an entirely new property: complexity (captured through \textit{sparsity}, \texttt{C} in Table~\ref{tab:properties}i).

% In the transductive setting, we have the ability to efficiently optimize a large number of additional formalizations of properties by recasting them as quadratic or linear programs. In particular, we will consider one additional formalization of faithfulness (\emph{function matching}), one additional formalization of robustness (\emph{maximum difference}) and a formalization of complexity (\emph{sparsity}). In this paper we do not derive instantiations of Equation~\ref{eqn:obj_trans} using these alternate formalizations of properties from the framework of GP regression. However, it would be interesting future work to consider kernels in the inductive setting such that they yield optimization problems in the transducitve setting with different formalization of properties.

\begin{itemize}
    \item \textit{Faithfulness as function matching.}
    This property is defined by a quadratic loss, $\sum\nolimits_n\|\*w_{E_n}-\nabla f(\*x_n)\|_2^2$, and can be optimized with an unconstrained quadratic program.
    
    \item \textit{Robustness as maximum difference.}
    The loss defining this property involves a $\+{max}$ operation: $\sum\nolimits_n\max\nolimits_{n'}\|\*w_{E_n}-\*w_{E_{n'}}\|_2^2$. Optimizing it can be recast as a linear program with quadratic constraints:
    \begin{align}
        \begin{split}
            \textit{minimize}& ~ {}_{W_E,\Delta_n}~~ \textstyle\sum_n \Delta_n \\[-1pt]
            \textit{s.t.}& ~ \|\*w_{E_n}-\*w_{E_{n'}}\|_2^2 \leq \Delta_n ~~\forall n,n'    
        \end{split}
        \label{eq:max_sens}
    \end{align}

% FDV: How exactly do we do it? Just for making this section have enough technical in it, is it easy to include the formula? 
    \item \textit{Complexity as sparsity.}
    Optimizing the loss defining this property, $\sum\nolimits_n\|W_{E_n}\|_1$ involves an unconstrained $\ell_1$ optimization, which can be rewritten as a linear program with linear constraints:
    \begin{alignat}{3}
            \textit{minimize}& ~{}_{W_E,\Delta_{nd}}~~ \textstyle\sum_{n,d} \Delta_{nd} \nonumber \\[-1pt]
            \textit{s.t.}& ~ \Delta_{nd} \geq 0 &&~~\forall n,d \\[-1pt]
            & ~ \Delta_{nd}\geq (W_{E_n})_d \geq -\Delta_{nd} &&~~\forall n,d \nonumber
    \end{alignat}
    When integrated with other objectives, it can be efficiently solved via sequential quadratic programs \citep{schmidt2005least}. 
\end{itemize}

In all of these optimizations, the trade-off parameters of our method, that is $\lambda_{\texttt{F-grad}}$, $\lambda_{\texttt{F-func}}$, $\lambda_{\texttt{R-avg}}$, $\lambda_{\texttt{R-max}}$, $\lambda_{\texttt{S}}$, and $\lambda_{\texttt{C}}$, allow us to explicitly and flexibly prioritize their corresponding properties. As we will show next with experiments, baselines like SmoothGrad, LIME, AGG, and MOFAE can only do so in limited ways, if at all.

%%%%%%%%%%%%%%%%%%%%%%%%%%%%%%%%%%%%%%%
%%%%%%%%%%%%%%%%%%%%%%%%%%%%%%%%%%%%%%%
%%%%%%%%%%%%%%%%%%%%%%%%%%%%%%%%%%%%%%%
%%%%% EXPERIMENTS
%%%%%%%%%%%%%%%%%%%%%%%%%%%%%%%%%%%%%%%
%%%%%%%%%%%%%%%%%%%%%%%%%%%%%%%%%%%%%%%
%%%%%%%%%%%%%%%%%%%%%%%%%%%%%%%%%%%%%%%

\section{Experiments}

First, we focus on the transductive setting (Section~\ref{sec:exp-transductive}, and demonstrate that our method can optimize for desired properties and manage the trade-offs between competing properties, while other baselines cannot reliably. We highlight how different formalizations of a property can be used in our method while still providing similar explanations across alternative formalizations.
Afterwards in Section~\ref{sec:exp-inductive}, we show that the same optimization can also be performed efficiently in the inductive setting, by selecting a small subset of the input over which to perform GP inference. Specifically, we check that the explanations generated for a test set inductively, based on a separate set of inducing points, approach those that could have been generated transductively if the test points were available ahead of time. This results means the two settings provide solutions that are consistent with each other, and the inductive setting enjoys the same benefits we highlight for the transductive setting. 
Our implementation is available at: \url{https://github.com/dtak/POE}

\subsection{Transductive Experiments}
\label{sec:exp-transductive}

\textbf{Baselines.}
We compare our approach to two popular explanation methods, SmoothGrad, and LIME as well as the two multi-property frameworks that we have discussed in our introduction, AGG and MOFAE:
\begin{itemize}
    \item \textit{SmoothGrad} is defined by averaging gradients of $f$ over a set of points sampled from a neighborhood of $\bm{x}_n$:
    \begin{align}
        \bm{w}_{\text{SG}_n} = \textstyle\frac{1}{S}\sum_{s=1}^S \nabla f(\tilde{\bm{x}}_{n,s}),~ \tilde{\bm{x}}_{n,s}\!\sim\!\mathcal{N}(\bm{x}_n,\delta_{\text{SG}}^2 I) \label{eqn:smoothgrad}
    \end{align}
    where the parameter $\delta_{\text{SG}}$ controls the size of the neighborhood, hence the degree of smoothing applied to $\nabla f$.
    
    \item \looseness-1\textit{LIME} approximates $f$ at $\bm{x}_n$ with a linear model trained on points $\tilde{\bm{x}}_{n,s}$ sampled from a $\delta_{\text{LIME}}$-neighborhood of $\bm{x}_n$:
    \begin{align}
        \bm{w}_{\text{LIME}_n} = \+{argmin}_{\bm{w}} \textstyle\sum_{s=1}^S (f(\tilde{\bm{x}}_{n,s})-\bm{w}^{\top}\tilde{\bm{x}}_{n,s})^2
    \end{align}
    
    \item \textit{AGG} linearly combines a base set of explanations $\{E^{(m)}\}$ with weights $\alpha_m$ to optimize for a desired property $\mathcal{L}$:
    \begin{align}
        \+{min}_{\begin{aligned}\scriptstyle \alpha_m: \Sigma_m\alpha_m&\scriptstyle=1 \\[-7pt] \scriptstyle\alpha_m&\scriptstyle\geq 0\\[-3pt]\end{aligned}}\: \mathcal{L}(W_{\text{AGG}} \doteq \textstyle\sum_m \alpha_m W_{E^{(m)}})
    \end{align}
    Like our approach, AGG can also manage trade-offs between different properties if we set $\mathcal{L}=\sum_i \lambda_{\text{prop}_i}\mathcal{L}_{\text{prop}_i}$. However, the range and the optimality of trade-offs it can achieve would naturally be limited by the initial properties of the base explanations~$\{E^{(m)}\}$. We initialize this set via SmoothGrad and LIME with varying hyperparameters (picking three explanations from each method).
    
    \item \textit{MOFAE} tackles multiple properties at once and searches for Pareto optimal explanations---those are explanations, $W_{\text{MOFAE}}$, which are better than any other explanation in terms of at least one property:
    \begin{align}
        \forall W_{E'},\:\: \exists i,\:\: \mathcal{L}_{\text{prop}_i}(W_{E'}) < \mathcal{L}_{\text{prop}_i}(W_{\text{MOFAE}})
    \end{align}
    However, MOFAE relies on a genetic algorithm to achieve this, and as such, returns random explanations along the Pareto front without any mechanism to control which properties are prioritized. We initialize MOFAE using the same base explanations as AGG but supplemented with 100 additional SmoothGrad explanations (as the genetic algorithm requires a significantly larger initial set).
\end{itemize}

\textbf{Functions.}
We consider a number of functions with different degrees of curvature, and periodicity/quasi-periodicity. Specifically, we consider:
\begin{itemize}
    \item \textit{Polynomials:} $f(\*x)=\sum_d(\*x)_d^3$ (\textit{Cubic}), $f(\*x)=\sum_d(\*x)_d^3$ $+\sum_{d,d',d''}(\*x)_d(\*x)_{d'}(\*x)_{d''}$ (\textit{Cubic with Interactions}).
    \item \textit{Periodic:} $f(\*x)=\sum_d\sin((\*x)_d)$.
    \item \textit{Quasi-Periodic:} We create quasi-periodic functions by modifying the above sinusoidal with various non-periodic terms, $f(\*x)=\sum_d(\*x)_d+\sin(3(\*x)_d)$ (\textit{with Linear Term}), $f(\*x)=\sum_d(\*x)_d^2/10+\sin(3(\*x)_d)$ (\textit{with Quadratic Term}), $f(\*x)=\sum_d\sin(e^{(\*x)_d/10})$ (\textit{with Exponentiated Inputs}), $f(\*x)=(\*x)_d+\sum_d\sin(e^{(\*x)_d/10})$ (\textit{with Linear Terms and Exponentiated Inputs}).
    \item \textit{Exponential:} $f(\*x)=\sum_d e^{(\*x)_d}$.
\end{itemize}

In addition to these simple functions, we also consider neural networks with one hidden layer in the appendix as well as pretrained convolutional neural networks for image classification later in Section~\ref{sec:exp-inductive} for the inductive setting. As query points, we choose $N=100^D$ points from a $D$-dimensional grid with values in $(-5,5)$ for each dimension. For our main experiments, we set $D=3$ but experiments for larger $D$ can be found in the appendix. For robustness losses, we consider a uniform similarity measure such that $S_{nn'}=1$. Finally, for our method, we vary the trade-off parameters $\lambda_{\text{prop}_i}$ so that they always sum to one.

\begin{figure}
    \centering
    \includegraphics[width=\linewidth]{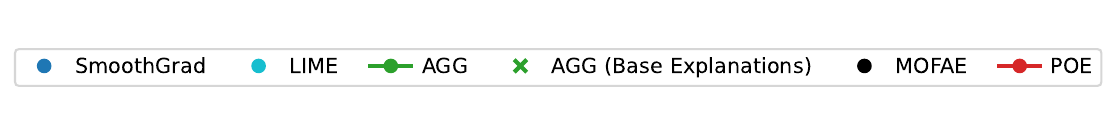} \\[3pt]
    \begin{minipage}{.05\linewidth}
        \rotatebox{90}{\tiny\sffamily \makecell{Faithfulness Loss\\(Gradient Matching, $\mathcal{L}_{\texttt{F-grad}}$)}}
    \end{minipage}%
    \begin{minipage}{0.99\linewidth}
        \begin{minipage}{.33\linewidth}
            \centering
            {\tiny\sffamily\textbf{\hspace{3pt}Cubic}} \\[-1pt]
            \includegraphics[width=\linewidth,trim={0 0 0 5pt},clip]{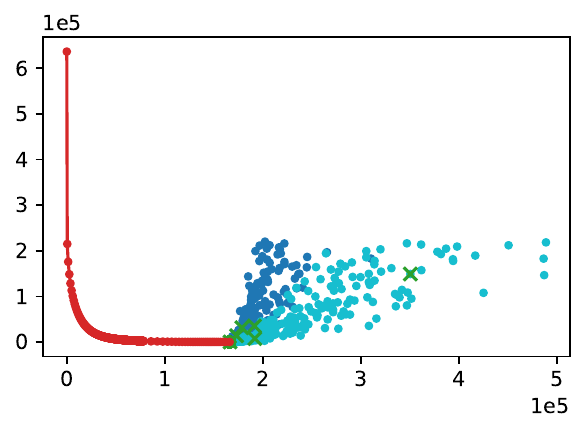}
        \end{minipage}%
        \begin{minipage}{.33\linewidth}
            \centering
            {\tiny\sffamily\textbf{\hspace{3pt}Exponential}} \\[-1pt]
            \includegraphics[width=\linewidth,trim={0 0 0 5pt},clip]{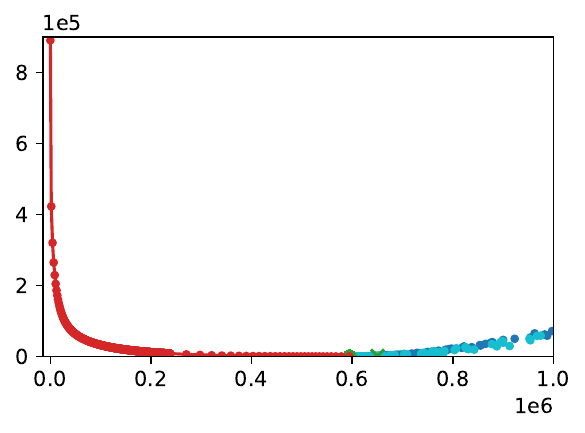}
        \end{minipage}%
        \begin{minipage}{.33\linewidth}
            \centering
            {\tiny\sffamily\textbf{\hspace{3pt}Periodic}} \\[-1pt]
            \includegraphics[width=\linewidth,trim={0 0 0 5pt},clip]{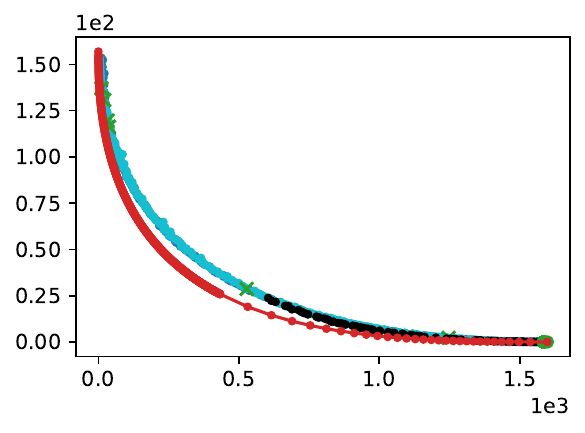}
        \end{minipage}\\
        \begin{minipage}{.33\linewidth}
            \centering
            {\tiny\sffamily\textbf{\hspace{3pt}Cubic\\[-\baselineskip]\hspace{3pt}with Interactions}} \\[-1pt]
            \includegraphics[width=\linewidth,trim={0 0 0 9pt},clip]{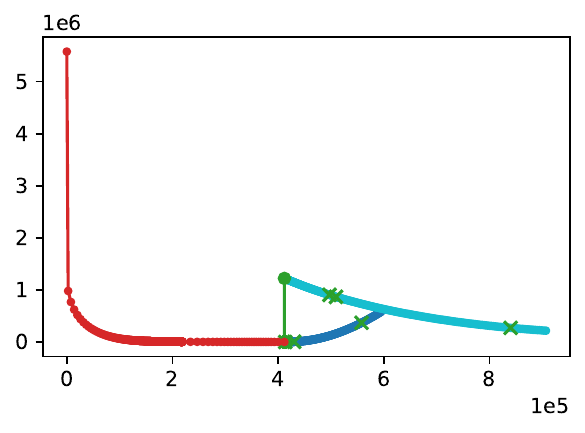}
        \end{minipage}%
        \begin{minipage}{.33\linewidth}
            \centering
            {\tiny\sffamily\textbf{\hspace{3pt}Quasi-Periodic\\[-\baselineskip]\hspace{3pt}with Exponentiated Inputs}}\\[-1pt]
            \includegraphics[width=\linewidth,trim={0 0 0 5pt},clip]{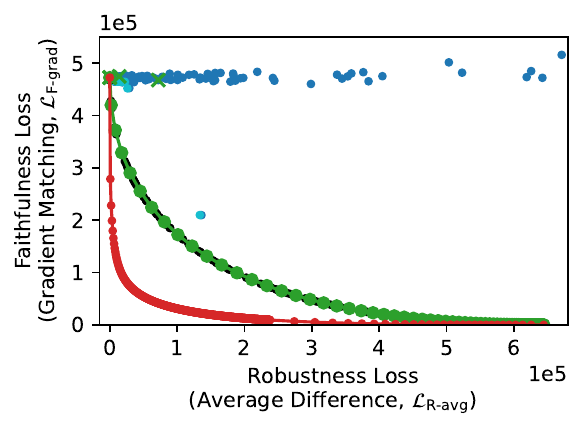}
        \end{minipage}%
        \begin{minipage}{.33\linewidth}
            \centering
            {\tiny\sffamily\textbf{\hspace{3pt}Quasi-Periodic\\[-\baselineskip]\hspace{3pt}with Quadratic Term}} \\[-1pt]
            \includegraphics[width=\linewidth,trim={0 0 0 5pt},clip]{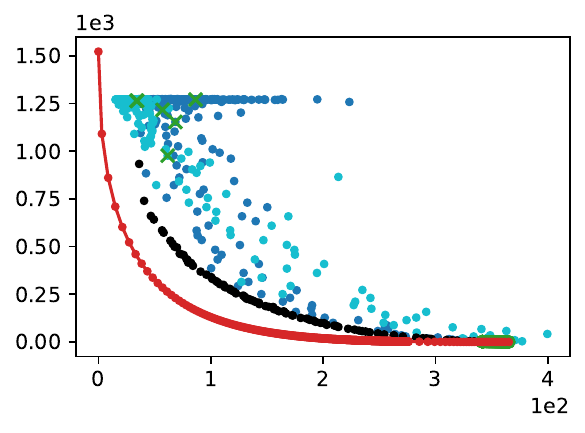}
        \end{minipage}
    \end{minipage}%

    % Hiwot: I will update the cubic interaction and the quadratic_quasi plots here. 
    
    {\tiny\sffamily{} Robustness Loss (Average Difference, $\mathcal{L}_{\texttt{R-avg}}$)}%
    \caption{\textbf{Faithfulness vs.\ Robustness.} For most functions, POE is the only method capable of generating explanations that cover the entire optimality front. SmoothGrad and LIME do not always result in a faithfulness-robustness balance, sometimes outputting strictly worse explanations when their hyperparameters are varied (e.g.\ \textit{Exponential}). AGG is constrained by the limited range of these methods. MOFAE fail to consistently cover the optimality front, often missing large portions (e.g.\ \textit{Exponential}) or even outputting suboptimal explanations (e.g.\ \textit{Cubic with Interactions}).}
    \label{fig:transductive-duo}
\end{figure}

\begin{figure}
\centering
    \includegraphics[width=\linewidth]{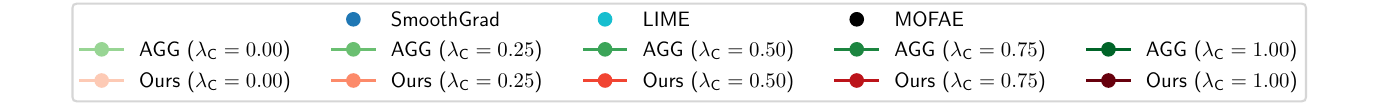} \\[3pt]
    \begin{minipage}{.08\linewidth}
        \rotatebox{90}{\tiny\sffamily \makecell{Faithfulness Loss\\(Gradient Matching)\\$\mathcal{L}_{\texttt{F-grad}}$}}
    \end{minipage}%
    \begin{minipage}{.92\linewidth}
        \begin{minipage}{.33\linewidth}
            \centering
            {\tiny\sffamily\textbf{\hspace{3pt}\makecell{\\Cubic}}} \\[-1pt]
            \includegraphics[width=\linewidth,trim={0 0 0 6pt},clip]{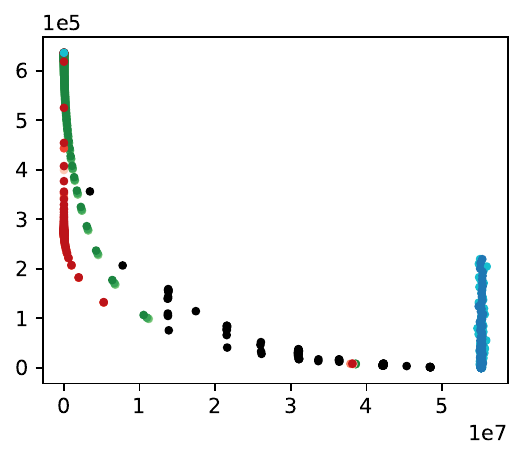}
        \end{minipage}%
        \begin{minipage}{.33\linewidth}
            \centering
            {\tiny\sffamily\textbf{\hspace{3pt}\makecell{Quasi-Periodic\\with Linear Term}}} \\[-2pt]
            \includegraphics[width=\linewidth,trim={0 0 0 6pt},clip]{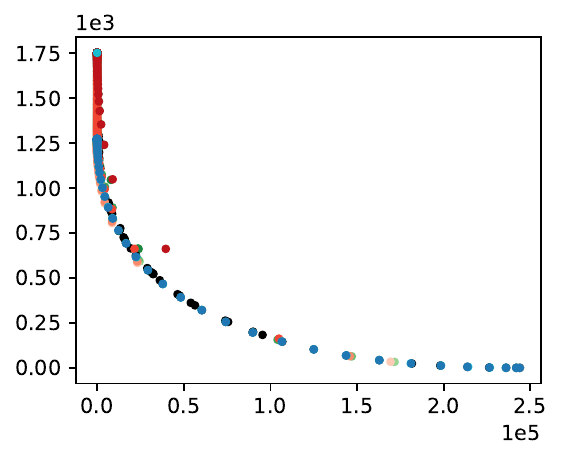}
        \end{minipage}%
        \begin{minipage}{.33\linewidth}
            \centering
            {\tiny\sffamily\textbf{\hspace{3pt}\makecell{Quasi-Periodic w/ Linear Term\\ \& Exponentiated Inputs}}} \\[-2pt]
            \includegraphics[width=\linewidth,trim={0 0 0 6pt},clip]{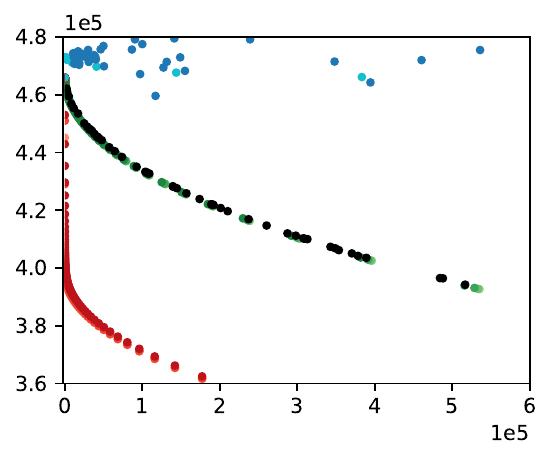}
        \end{minipage}
    \end{minipage}
    {\tiny\sffamily{} Robustness Loss (Max  Difference, $\mathcal{L}_{\texttt{R-max}}$)} \\[6pt]
    \includegraphics[width=\linewidth]{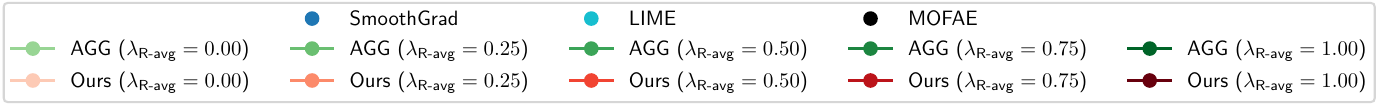} \\[3pt]
    \begin{minipage}{.08\linewidth}
        \rotatebox{90}{\tiny\sffamily \makecell{\\\\Complexity Loss ($\mathcal{L}_{\texttt{C}}$)}}
    \end{minipage}%
    \begin{minipage}{.92\linewidth}
        \begin{minipage}{.33\linewidth}
            \centering
            {\tiny\sffamily\textbf{\hspace{3pt}\makecell{\\Cubic}}} \\[-2pt]
            \includegraphics[width=\linewidth,trim={0 0 0 6pt},clip]{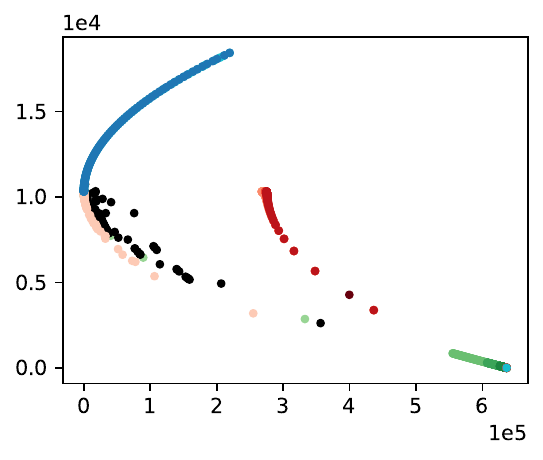}
        \end{minipage}%
        \begin{minipage}{.33\linewidth}
            \centering
            {\tiny\sffamily\textbf{\hspace{3pt}\makecell{Quasi-Periodic\\with Linear Term}}} \\[-2pt]
            \includegraphics[width=\linewidth,trim={0 0 0 6pt},clip]{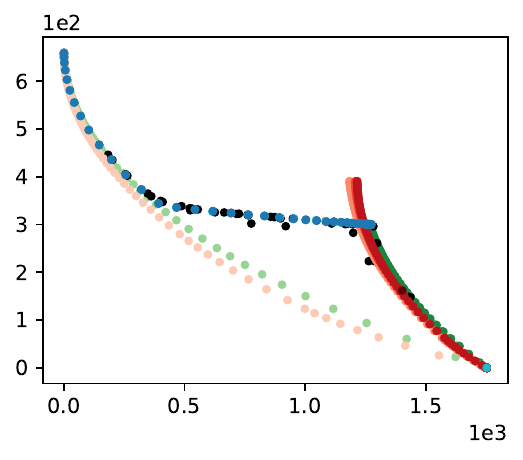}
        \end{minipage}%
        \begin{minipage}{.33\linewidth}
            \centering
            {\tiny\sffamily\textbf{\hspace{3pt}\makecell{Quasi-Periodic w/ Linear Term\\ \& Exponentiated Inputs}}} \\[-2pt]
            \includegraphics[width=\linewidth,trim={0 0 0 6pt},clip]{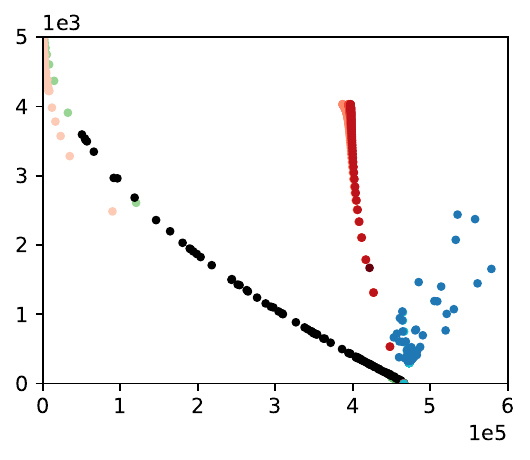}
        \end{minipage}
    \end{minipage}%
    {\tiny\sffamily{} Faithfulness Loss (Gradient Matching, $\mathcal{L}_{\texttt{F-grad}}$)}%
    \caption{\textbf{Faithfulness vs.\ Robustness vs.\ Complexity.} In the top row, we report faithfulness vs.\ robustness for fixed values of $\lambda_{\texttt{C}}$. In the bottom row, we report complexity vs.\ faithfulness for fixed values of $\lambda_{\texttt{R-max}}$.}
    \label{fig:transductive-trios}

\end{figure}

\textbf{Results.} In Figure~\ref{fig:transductive-duo}, we optimize for faithfulness (gradient matching) and robustness (average difference), varying $\lambda$ for us and AGG. We also run SmoothGrad and LIME, varying~$\delta$. In Figure~\ref{fig:transductive-trios}, we consider complexity in addition to faithfulness and robustness.

\textit{Intuitive interpretations of the hyperparameters of SmoothGrad and LIME do not always hold.} A number of works show that $\delta$, the sampling hyper-parameter of LIME and SmoothGrad, affect the robustness of these methods. In particular, when the $\delta$ is larger, LIME and SmoothGrad generate explanations that are less sensitive to local perturbations. However, robustness in these analyses is formalized as local-Lipschitzness, and it is not clear how their insight generalize to other formalizations. For instance, in Figure~\ref{fig:transductive-duo} for \textit{Cubic}, we see that the robustness of SmoothGrad (as defined in Equation~\ref{eq:robustness}) does not depend on $\delta$ at all---increasing $\delta$ does not smooth the explanations, only reduces their faithfulness.

\textit{Without explicit optimization, SmoothGrad and LIME can fail to find robust and faithful explanations.} In Figure~\ref{fig:transductive-duo} for \textit{Exponential} and \textit{Quasi-Periodic with Exponentiated Inputs}, we see that varying $\delta$ not only fails to control the balance between faithfulness and robustness, but also might lead to strictly worse explanations in terms of both properties.

\looseness-1
\textit{AGG is limited by the quality of its base explanations.} Although AGG, similar to our approach, allows transparent control over different properties, the optimality of the trade-offs it can achieve is constrained by the properties of its base explanations. For instance, in Figure~\ref{fig:transductive-duo} for \textit{Cubic}, AGG produces suboptimal explanations compared to POE, as the base explanations themselves are already suboptimal.

\looseness-1
\textit{MOFAE does not offer any control over trade-offs and might not always cover the complete range of possible trade-offs.} Since it uses a genetic algorithm to search for explanations, MOFAE essentially produces a random collection of explanations without offering any control over where these explanations land in terms of their properties. Sometimes, they do not achieve optimal trade-offs at all (e.g.\ Figure~\ref{fig:transductive-duo}, \textit{Cubic with Interactions}). At other times, they only offer a limited range of trade-offs, leaving portions of the Pareto front sparsely populated (e.g.\ Figure~\ref{fig:transductive-duo}, \textit{Cubic} and \textit{Exponential}, where explanations with low robustness loss are missed). For a zoomed-in version of the figure focusing on AGG and MOFAE, please refer to Appendix ~\ref{subsec:zoomed_trade-off}.

\looseness-1
\textit{Transparent Trade-offs between Properties.}
In contrast to all our baselines, POE can consistently find optimal trade-offs between different properties, but even more importantly, it provides fine control over these trade-offs through hyperparameters $\lambda_{\text{prop}_i}$. POE makes it significantly easier to include additional properties to the optimization as well, see results for complexity in Figure~\ref{fig:transductive-trios}. The addition of this third property does not affect the explanations generated by SmoothGrad and LIME since they do not optimize for any property. Similarly, the base explanation set of AGG does not get any richer either. It increases the dimensionality of the optimality front that MOFAE needs to cover, which makes its coverage even worse. For instance, we see in Figure~\ref{fig:transductive-trios} for \textit{Quasi-Periodic with Exponentiated Inputs}, MOFAE covers the faithfulness-robustness front well (top row) but misses a large portion of solutions that have low complexity loss (bottom row). Meanwhile, our approach with $\lambda_{\texttt{R-max}}=0$ covers the faithfulness-complexity front as well.

\subsection{Inductive Experiments}
\label{sec:exp-inductive}

\begin{figure}
    \centering
    \begin{minipage}{.03\linewidth}
        \rotatebox{90}{\tiny\sffamily{} Loss (Total, $\sum_i\lambda_{\text{prop}_i}\mathcal{L}_{\text{prop}_i}$)}
    \end{minipage}%
    \begin{minipage}{.97\linewidth}
        \begin{minipage}{.33\linewidth}
            \centering
            {\tiny\sffamily\textbf{\hspace{3pt}Cubic}} \\[-1pt]
            \includegraphics[width=\linewidth,trim={0 0 0 9pt},clip]{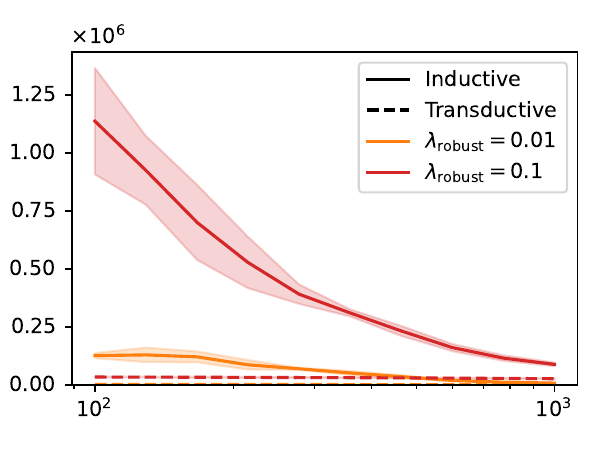}
        \end{minipage}%
        \begin{minipage}{.33\linewidth}
            \centering
            {\tiny\sffamily\textbf{\hspace{3pt}Exponential}} \\[-1pt]
            \includegraphics[width=\linewidth,trim={0 0 0 9pt},clip]{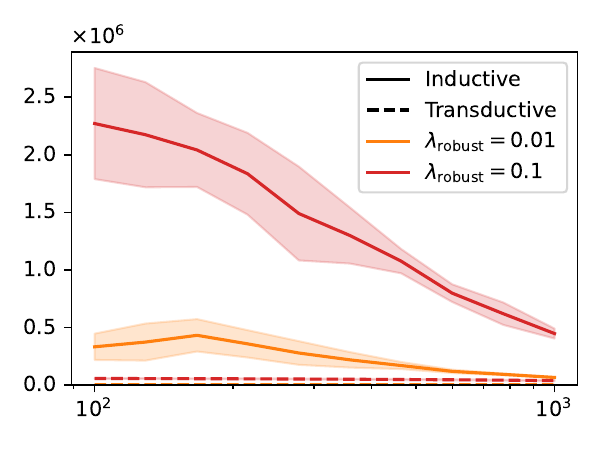}
        \end{minipage}%
        \begin{minipage}{.33\linewidth}
            \centering
            {\tiny\sffamily\textbf{\hspace{3pt}Periodic}} \\[-1pt]
            \includegraphics[width=\linewidth,trim={0 0 0 9pt},clip]{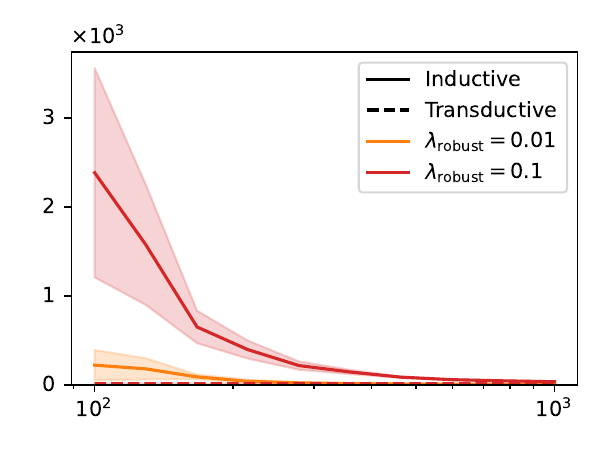}
        \end{minipage}\\
        \begin{minipage}{.33\linewidth}
            \centering
            {\tiny\sffamily\textbf{\hspace{3pt}Quasi-Periodic\\[-\baselineskip]\hspace{3pt}with Exponentiated Inputs}} \\[-1pt]
            \includegraphics[width=\linewidth,trim={0 0 0 9pt},clip]{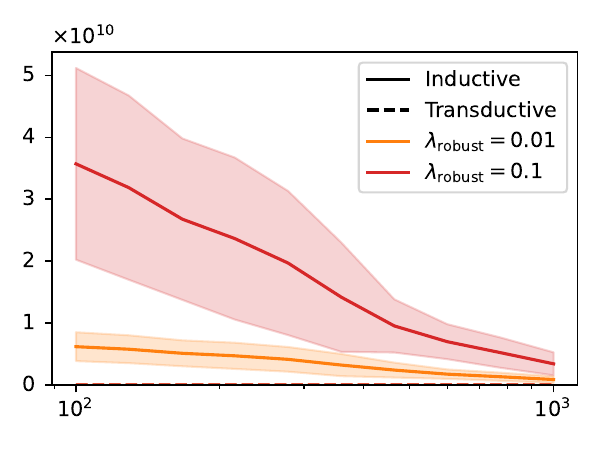}
        \end{minipage}%
        \begin{minipage}{.33\linewidth}
            \centering
            {\tiny\sffamily\textbf{\hspace{3pt}Quasi-Periodic\\[-\baselineskip]\hspace{3pt}with Linear Term}} \\[-1pt]
            \includegraphics[width=\linewidth,trim={0 0 0 9pt},clip]{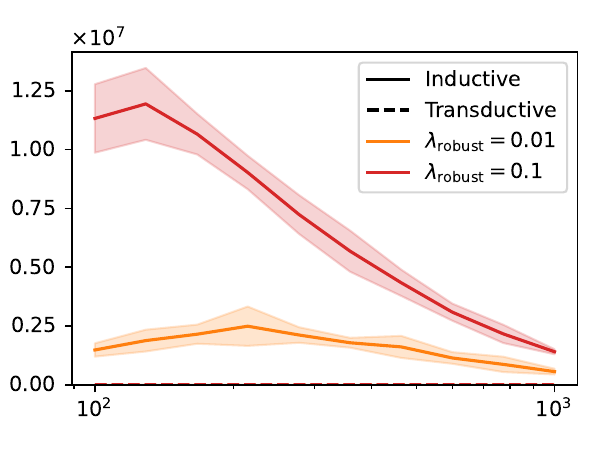}
        \end{minipage}%
        \begin{minipage}{.33\linewidth}
            \centering
            {\tiny\sffamily\textbf{\hspace{3pt}Pretrained CNN\\[-\baselineskip]\hspace{3pt}with ImageNet-1k}} \\[-1pt]
            \includegraphics[width=\linewidth,trim={0 0 0 9pt},clip]{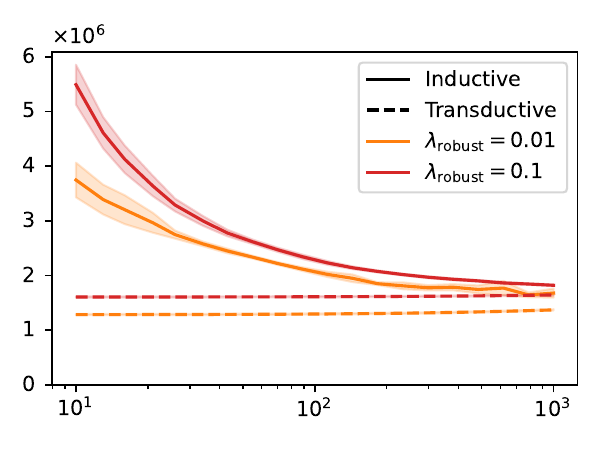}
        \end{minipage}
    \end{minipage}%
    {\tiny\sffamily{} Number of Inducing Points ($N$)}%
    \caption{As the number of inducing points increases, inductive solutions obtained via GP inference (optimizing faithfulness and robustness) approach to transductive solutions on an exponential scale.}
    \label{fig:inductive}
\end{figure}

\textbf{Quantitative Evaluation.} Our goal is to check the consistency between the transductive and inductive settings. We optimize for faithfulness (gradient-matching) and robustness (average differences), fixing $\lambda_{\texttt{F-grad}}=1$ and varying $\lambda_{\texttt{R-avg}}$. We report the total loss as we vary the number of inducing points $N$ that are used for GP inference, comparing this with the total loss achievable transductively. We investigate the impact of how the inducing points are selected in the appendix. For now, just like in Proposition~\ref{prop}, we assume that they share the same distribution as our query points.

In addition to functions from Section~\ref{sec:exp-transductive}, we also consider a \textit{Convolution Neural Network (CNN)}, specifically the model architecture ResNet-50 \citep{he2016deep} pretrained on the dataset ImageNet-1k \citep{russakovsky2015imagenet}. For quantitative results, we sample $1000$ query images from ImageNet-1k and up to $N=1000$ separate images to be used as inducing points. Note that, for these images, $D=244\times 244$. For the robustness loss, we consider similarity measures obtained by inverting Guassian kernels, $s=k^{-1}$ and $k(\*x,\*x')=\exp(-0.5\|\*x-\*x'\|_2^2/\Lambda^2)$.

In Figure~\ref{fig:inductive}, we show that inductive solutions approach transductive solutions exponentially fast as the number of inducing points are increased. This means the benefits we have seen so far can also be achieved in the inductive setting (although for a more restricted set of properties).

\textbf{Qualitative Evaluation.} Finally, we also provide a visual example to highlight the control over different properties provided by POE. We pick a single image from ImageNet-1k (one more example in the appendix) and compare the explanations generated by our approach against SmoothGrad. Using POE, we optimize for faithfulness, robustness, and smoothness via GP inference, using perturbations of the original image as inducing points (uniform perturbations as opposed to Gaussian perturbations). We fix $\lambda_{\texttt{F-grad}}=1$ and vary $\lambda_{\texttt{R-avg}}$ and $\lambda_{\texttt{S}}$. For SmoothGrad, we can only vary $\delta$ in Equation~\ref{eqn:smoothgrad} (using Gaussian perturbations to match the Gaussian kernel in our approach).

\begin{figure}
    \centering
    \includegraphics[width=\linewidth]{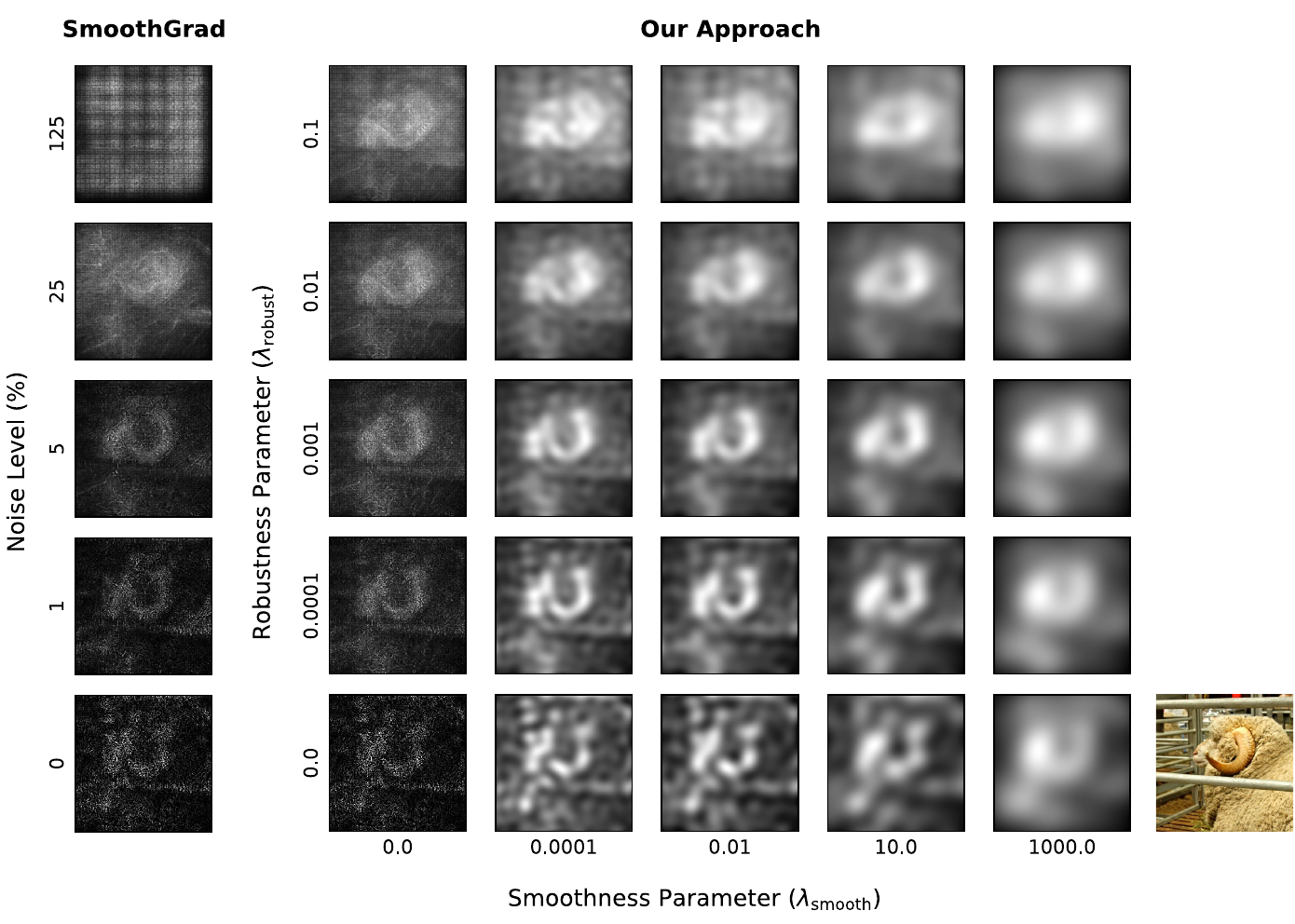}
    \caption{\justify{Our approach can achieve an arbitrary balance between faithfulness, robustness, and smoothness. Meanwhile, SmoothGrad can only manage the trade-off between faithfulness and robustness, and does not improve smoothness.}}
    \label{fig:qualitative-example0}
\end{figure}

In Figure~\ref{fig:qualitative-example0}, we see that POE can achieve an arbitrary balance between faithfulness, robustness, and smoothness. Furthermore, the range of explanations we obtain shows that robustness and smoothness each capture distinct aspects of what makes an explanation ``noisy''. While the main motivation behind SmoothGrad is to de-noise explanations, it attributes noisiness solely to sharp variations in the gradients around an input point, thereby only improving robustness while missing the smoothness aspect entirely. Meanwhile, users seeking de-noised explanations might have different preferences regarding which aspect to prioritize. Without a precise definition of what constitutes a noisy explanation, and without an explicit optimization targeting that definition, it is not clear how to modify SmoothGrad to also promote smoothness alongside robustness, whereas in our approach, noisiness can be expressed as a specific mixture of robustness and smoothness based on user needs.
\vspace{-1em}
\section{Ethical and Societal Implications}

As argued by \citet{alpsancar2024explanation}, explanations are a means to an end, they are valuable not for their own sake, but for how they support broader goals such as fairness, safety, accountability and trust in machine learning systems. Our work is intended to help users tailor explanations to serve these goals more effectively. We believe that our approach can give end users more control over how explanations align with their specific needs. For example, in a recent work \citet{nofshin2024sim2real} showed that when attempting to detect bias in a model, a less compact but more faithful explanation is essential. With our approach, the properties of an explanation – how faithful, how compact, etc– are transparent. This provides a means for users to know what the explanation can and cannot be used for, whether that is to help mitigating bias and supporting fairness in real-world applications, safety, or any other task. 

%Hiwot: Add more stuff here, also adress the Goodhart's law here 
\vspace{-1em}
\section{Conclusion and Future Work}
A growing number of studies show that different downstream applications may require explanations with different properties. Yet existing feature attribution methods like SmoothGrad and LIME neither directly optimize for arbitrary explanation properties, nor can they manage trade-offs between properties that are in tension. Multi-property frameworks like AGG and MOFAE are either constrained by the limitations of these methods or do not offer mechanisms to control trade-offs when optimizing for properties.

In our work, we introduced POE, a framework that can efficiently optimize for a set of desired explanation properties. In addition, POE can also explicitly and intuitively manage trade-offs between competing properties through its hyperparameters. We addressed both settings in which we have the entire dataset that needs explanations (the transductive setting) as well as settings in which we must generalize our property-optimized explanations to unseen query points (the inductive setting).

There are numerous potentially useful explanation properties. Some of those could use the same ideas in our work straightforwardly; others will require algorithmic innovation. Developing such methods is an interesting direction for future work.

Another limitation of our framework, shared with many feature-attribution explanation methods, is scalability. While POE is faster than existing methods for some properties, it becomes slower when solving a quadratic program is required. We see efficiently optimizing for a range of properties e.g. with clever amortization schemes as a ripe direction for future work. 

% There are a very large number of potentially useful properties.  Some of those could use the same ideas in our work straightforwardly; others will require more algorithmic innovation. More broadly, for the optimization itself, there are interesting future directions in scalability – while for certain properties, our approach is actually also faster than existing approaches, when a quadratic program is required, it is slower.  We see efficiently optimizing for a range of properties e.g. with clever amortization schemes as a ripe direction for many future works. 

% \begin{contributions}
% \end{contributions}

\begin{acknowledgements}
This material is based upon work supported by the National Science Foundation under Grant No. IIS-2007076. Any opinions, findings, and conclusions or recommendations expressed in this material are those of the author(s) and do not necessarily reflect the views of the National Science Foundation.

This material is based upon work supported by the National Science Foundation under Grant No. IIS-1750358. Any opinions, findings, and conclusions or recommendations expressed in this material are those of the author(s) and do not necessarily reflect the views of the National Science Foundation

Supported by NIH award 5R01MH123804-02.
\end{acknowledgements}

% references
\bibliography{uai2025-template}

\newpage
\onecolumn

\title{Transparent Trade-offs between Properties of Explanations\\(Supplementary Material)}
\maketitle
\appendix
\section{Choice of Similarity Function}
For robustness losses, we need to choose a similarity function $s(\*x,\*x')$. Certain similarity functions lend themselves to more efficient optimization. 
In particular, in Section~\ref{sec:inductive}, we showed that when $s(\*x,\*x')=-k^{-1}(\*x,\*x')$ is the precision of a kernel function $k$, optimization in the inductive setting becomes equivalent to Gaussian process inference. 
As an example, in this work, we consider similarity measures that are induced by Gaussian kernels: $k(\*x,\*x';\Lambda) = \exp(-\|\*x-\*x'\|_2^2 / 2\Lambda^2)$. 
%
% (i) $k_{\text{threshold}}(\*x,\*x';\Lambda)=\mathbbm{1}\{\|\*x - \*x'\|_2^2 \leq \Lambda\}$ and (ii)~$k_{\text{gaussian}}(\*x,\*x';\Lambda) = \exp\left\{-\|\*x-\*x'\|_2^2 / 2\Lambda^2\right\}$. 
%
In practice, we note that the choice of $s$ or an associated kernel $k$ should be informed by domain knowledge (e.g.\ capture a notion of similarity between inputs that is meaningful for the user's downstream task).

\section{Proof of Proposition~\ref{prop}}

For a sufficiently large $N$, the objective in Equation~\ref{eqn:prop-objective} can be written as:
\begin{align}
    \+{argmin}_{E} \frac{1}{N}\sum\nolimits_n\|E(\*x_n)-\nabla f(\*x_n)\|_2^2 - \frac{1}{N^2}\sum\nolimits_{n,n,d,d'} |E_d(\*x_n)-E_{d'}(\*x_{n'})|^2 K^{-1}_{dd'}(\*x_n,\*x_n') \label{eqn:proof-objective}
\end{align}
We show that, as $N\to\infty$ and $\sigma^2\to\infty$, maximizing $\log p(E|\{\*x_n,\nabla f(\*x_n)\})$ becomes equivalent to this objective.

For simplicity, we sitch to a vectorized notation so that we can work with a single-dimensional GP: $\*E = [E(\*x_1) \ldots E(\*x_N)] \in \mathbb{R}^{ND}$, $\*{\nabla f} = [\nabla f(\*x_1) \ldots \nabla f(\*x_N)] \in \mathbb{R}^{ND}$, and 
\begin{equation}
    \*K = \begin{bmatrix}
        K(\*x_1,\*x_1) & \ldots & K(\*x_1,\*x_N) \\
        \vdots & \ddots & \vdots \\
        K(\*x_N,\*x_1) & \ldots & K(\*x_N,\*x_N)
    \end{bmatrix} \in \mathbb{R}^{ND\times ND}
\end{equation}

Let us first consider only the prior term in the GP, marginalizing out the random mean explanation $\mu$ from the joint distribution:
\begin{align}
    p(\*E, \mu) = p(\*E|\mu) p(\mu) \propto \exp\left \{ -\frac{1}{2} (\*E-\mu\*1_{ND})^\top \*K^{-1} (\*E-\mu\*1_{ND})\right \} \exp\left \{ -\frac{1}{2} \cdot \frac{\mu^2}{\sigma^2} \right \} \label{eqn:joint}
\end{align}
where $\*1_{ND}\in\mathbb{R}^{ND}$ is a vector of ones (written as $\*1$ henceforth).
By marginalizing out $\mu$ in Equation \ref{eqn:joint} (after completing the square, followed by some algebra), we obtain $p(E)$ as the following Gaussian:
\begin{equation}
p(\*E) \propto \exp \bigg\{ -\frac{1}{2} \*E^\top\bigg( \underbrace{\*K^{-1} - \frac{\*K^{-1} \*{1} \*{1}^\top \*K^{-1}}{\*{1}^\top\*K^{-1}\*{1} + 1/\sigma^2}}_{\doteq \Sigma^{-1}} \bigg )\*E\bigg\}
\label{eqn:prior_cov}
\end{equation}
We denote the covariance of $p(\*E)$ by $\Sigma$. This marginalization can be interpreted as a new Gaussian prior on the explanations $\*E$; we will still have a multivariate Gaussian---and associated GP---with this new kernel.

Next, we note that the posterior mean of multivariate Gaussian $p(\*E|\*{\nabla f})$ is the solution to the least square regression problem, regularized by the precision matrix $\Sigma^{-1}$ of the prior $p(\*E)$ that we just derived above in Equation~\ref{eqn:prior_cov}:
\begin{align}
    \max\nolimits_{\*E} \log p(\*E|\*{\nabla f}) &\equiv \min\nolimits_{\*E} \frac{1}{\varsigma^2} (\*E-\*{\nabla f})^\top (\*E-\*{\nabla f}) + \*E^\top \Sigma^{-1} \*E \\
    &\equiv \min\nolimits_{\*E} \frac{1}{N} (\*E-\*{\nabla f})^\top (\*E-\*{\nabla f}) + \frac{1}{N^2} \*E^\top \Sigma^{-1} \*E \label{eqn:gp_obj}
\end{align}
With a little algebraic manipulation, the regularization term $E^\top \Sigma^{-1} E$ can be further expanded as follows:
\begin{align}
\label{eqn:reg}
\begin{split}
\*E^\top \Sigma^{-1} \*E =&\underbrace{-\frac{1}{2} \sum_{\ell=1}^{ND}\sum_{\ell'=1}^{ND} |\*E_{\ell}-\*E_{\ell'}|^2\Sigma^{-1}_{\ell\ell'}}_{\text{Term 1}} + \underbrace{\sum_{\ell=1}^{ND} \left(|\*E_{\ell}|^2 \sum_{\ell'=1}^{ND} \Sigma^{-1}_{\ell\ell'}\right)}_{\text{Term 2}}
\end{split}
\end{align}

We first observe that the second term goes to zero as the variance of the explanation mean $\sigma^2$ goes to infinity.  Specifically, Term 2 in Equation \ref{eqn:reg} can be expanded as:
\begin{equation}
    \sum_\ell |\*E_{\ell}|^2 \left( \delta_{\ell}^{\top}\*K_{\ell}^{-1}\*{1}  - \delta_{\ell}^{\top}\*K_{\ell}^{-1} \*1\left(\frac{\*1^\top \*K^{-1}\*1}{\*{1}^\top\*K^{-1}\*{1} + 1/\sigma^2} \right)\right)
\end{equation}
where $\delta_{\ell}$ is a vector of zeros except for the $\ell$-th element, which is one. As $\sigma^2$ goes to infinity (representing an improper, uninformative prior over the value of the mean explanation), the fraction goes to one and the first and second terms in the kernel expression cancel.

Thus, we are left with only the first term:
\begin{align}
    \*E^\top \Sigma^{-1} \*E \approx &\underbrace{-\frac{1}{2} \sum_{\ell=1}^{ND}\sum_{\ell'=1}^{ND} |\*E_{\ell}-\*E_{\ell'}|^2\Sigma^{-1}_{\ell\ell'}}_{\text{Term 1}}
\end{align}

We are going to finish our proof by showing that $\Sigma^{-1}$ tends to $\bm{K}^{-1}$ as $\sigma^2\to\infty$ and $N\to\infty$. For that, we rely on the fact that, as $N$ gets larger, the normalized kernel $\bm{K}/N$ becomes a better approximation of the operator
\begin{align}
    (Th)_d(\*x) = \mathbb{E}_{\bm{x}'\sim\mathcal{P}}\left[\sum\nolimits_{d'} K_{dd'}(\*x,\*x') h_{d'}(\*x')\right]
\end{align}
over the vector-valued functions $h:\mathbb{R}^D\to\mathbb{R}^D$. More precisely,
\begin{align}
    (\bm{K}/N)[h(\*x_1)\dots h(\*x_N)] \to [(Th)(\*x_1)\dots (Th)(\*x_N)]
\end{align}
as $N\to\infty$. Hence, given $g$ such that $(Tg)(\*x)=\bm{1}$ for all $\*x\in\mathbb{R}^D$, we have $N\*{K}^{-1}\bm{1}\to [g(\*x_1)\ldots g(\*x_N)]$. This has two implications that are important to us:
\begin{align}
    \|(N\*{K}^{-1}\bm{1})\|_{\infty} &\leq \max\nolimits_{\*x}\|g(\*x)\|_{\infty} \\
    \*1^\top(\*K^{-1}\*1) &= \frac{\*1^{\top}}{N}(N\*K^{-1}\*1) = \mathbb{E}_{\*x\sim\mathcal{P}}\left[\sum\nolimits_{d}g_d(\*x)\right]
\end{align}

With those facts, we are ready to show the convergence of $\Sigma^{-1}$:
\begin{align}
    \Sigma^{-1}_{\ell\ell'} &= \*K_{\ell\ell'}^{-1} -\frac{\delta_{\ell}^{\top}(\*K^{-1}\*1)(\*K^{-1}\*1)^{\top}\delta_{\ell'}}{\*1^\top(\*K^{-1}\*1) + 1/\sigma^2} \to \*K_{\ell\ell'}^{-1}
\end{align}
as $\sigma^2\to\infty$ and $N\to\infty$ since
\begin{align}
    \frac{\delta_{\ell}^{\top}(\*K^{-1}\*1)(\*K^{-1}\*1)^{\top}\delta_{\ell'}}{\*1^\top(\*K^{-1}\*1) + 1/\sigma^2} &\leq \frac{1}{N^2} \cdot \frac{\|N\*K^{-1}\*1\|_{\infty}^2}{\*1^\top(N\*K^{-1}\*1) + 1/\sigma^2} = \frac{1}{N^2}\cdot\frac{\max\nolimits_{\*x}\|g(\*x)\|_{\infty}}{\mathbb{E}_{\*x\sim\mathcal{P}}\left[\sum\nolimits_{d}g_d(\*x)\right]+1/\sigma^2} \to 0
\end{align}
\vspace{-1.5em}

\section{Additional Transductive Setting Results}

\subsection{Closer Look at AGG and MOFAE: Faithfulness vs. Robustness}
\label{subsec:zoomed_trade-off}
In this section, we provide a detailed comparison of POE with AGG and MOFAE by zooming into the region of interest in Figure~\ref{fig:transductive-duo}. The right panel displays all baselines, with a highlighted rectangle indicating the area magnified for analysis. This zoomed-in view isolates the behavior of AGG and MOFAE, illustrating their suboptimal solution in contrast to POE.
\begin{figure}[H]
\includegraphics[width=\linewidth]{figures/duo-legend1} 
\vspace{-2pt}
    \centering
    \includegraphics[width=\linewidth]{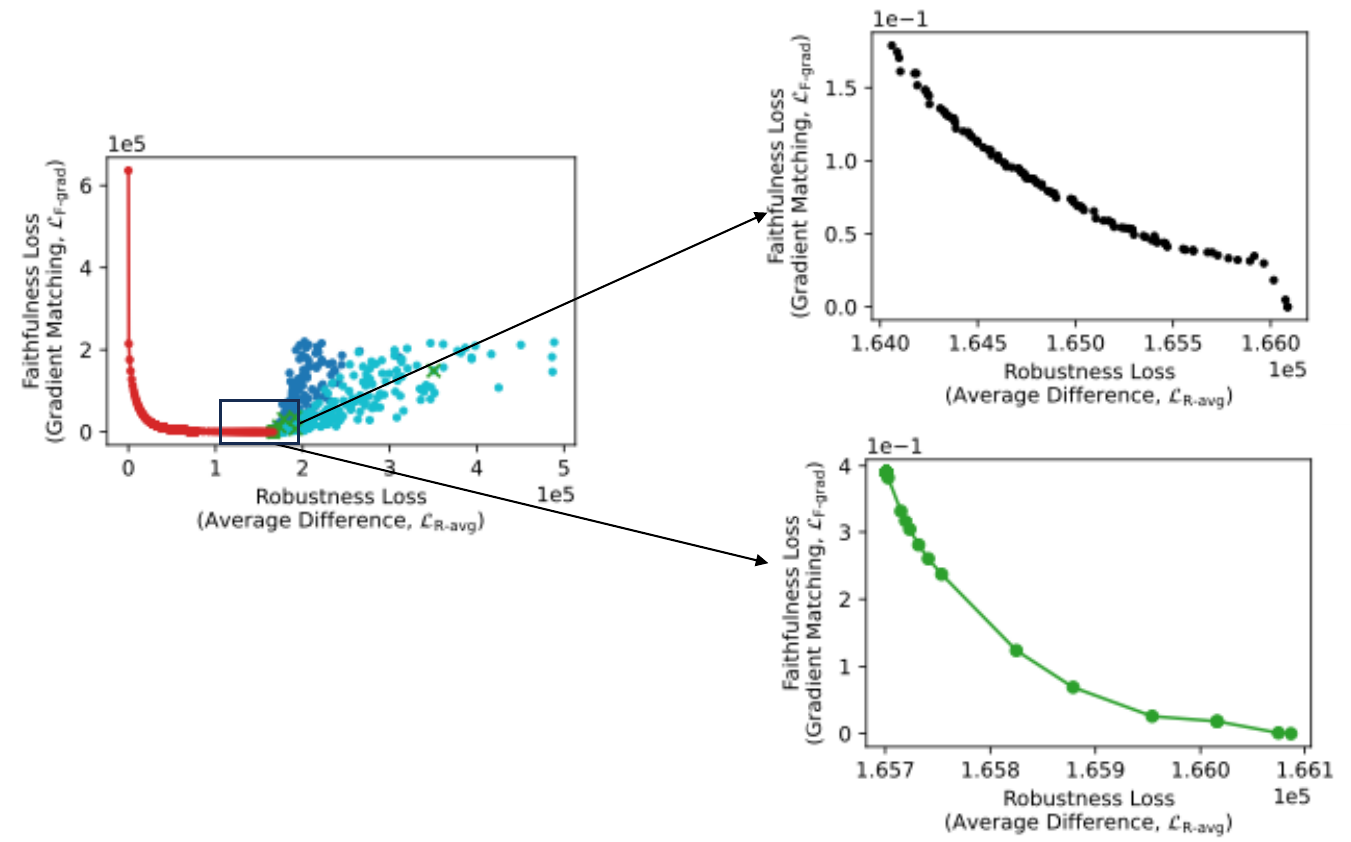}
    \caption{Comparison of POE with AGG and MOFAE for a cubic function}
    \label{fig:enter-label}
\end{figure}
\vspace{-10em}
\subsection{SHAP}

SHAP \cite{lundberg2017unified} generates explanations by sampling subsets of features to quantify their contributions. When comparing SHAP’s faithfulness and robustness losses to other explanation methods—including our proposed method, POE we observe distinct behavior which for some functions the SHAP explanations have a higher robustness and faithfulness loss. One potential reason for this difference could be SHAP’s lack of tunable hyperparameters, which sets it apart from alternative methods.  We plan to explore this in our future work.

% \begin{figure}[H]
%     \centering
%     \includegraphics[width=\linewidth]{figures-w-kernel/shap-legend} \\[3pt]

% \end{figure}

\begin{figure}[H]
    \centering
    \includegraphics[width=\linewidth]{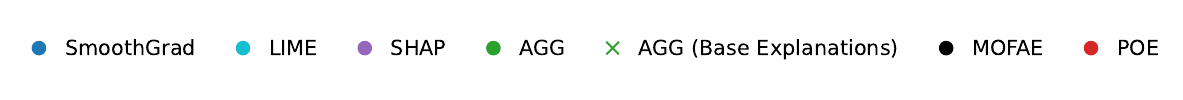} \\[3pt]
    \begin{minipage}{.05\linewidth}
        \rotatebox{90}{\tiny\sffamily \makecell{Faithfulness Loss\\(Gradient Matching, $\mathcal{L}_{\texttt{F-grad}}$)}}
    \end{minipage}%
    \begin{minipage}{.95\linewidth}
        \begin{minipage}{.33\linewidth}
            \centering
            {\tiny\sffamily\textbf{\hspace{3pt}Cubic}} \\[-1pt]
            \includegraphics[width=\linewidth,trim={0 0 0 9pt},clip]{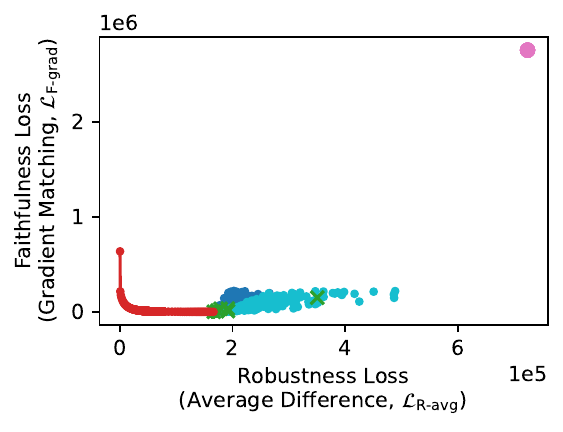}
        \end{minipage}%
        \begin{minipage}{.33\linewidth}
            \centering
            {\tiny\sffamily\textbf{\hspace{3pt}Exponential}} \\[-1pt]
            \includegraphics[width=\linewidth,trim={0 0 0 9pt},clip]{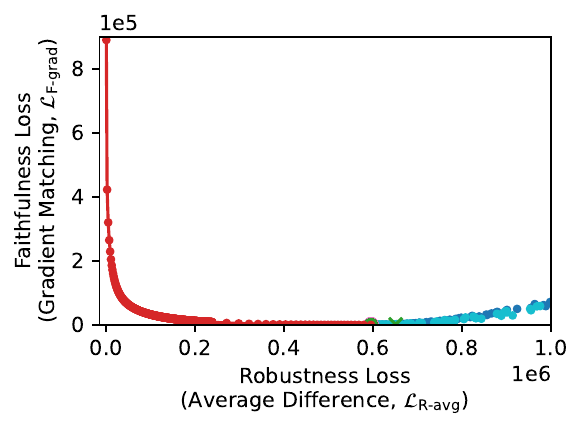}
        \end{minipage}%
        \begin{minipage}{.33\linewidth}
            \centering
            {\tiny\sffamily\textbf{\hspace{3pt}Periodic}} \\[-1pt]
            \includegraphics[width=\linewidth,trim={0 0 0 9pt},clip]{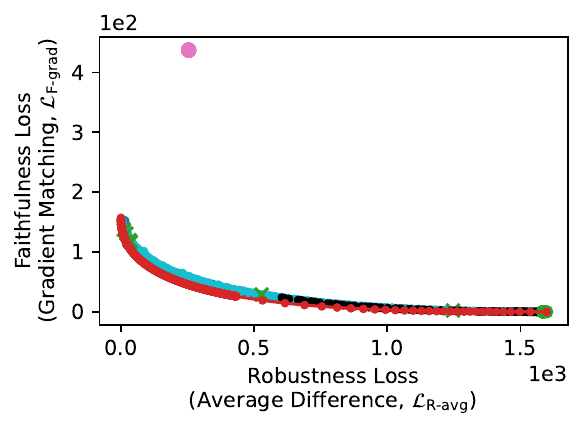}
        \end{minipage}\\
        \begin{minipage}{.33\linewidth}
            \centering
            {\tiny\sffamily\textbf{\hspace{3pt}Cubic\\[-\baselineskip]\hspace{3pt}with Interactions}} \\[-1pt]
            \includegraphics[width=\linewidth,trim={0 0 0 9pt},clip]{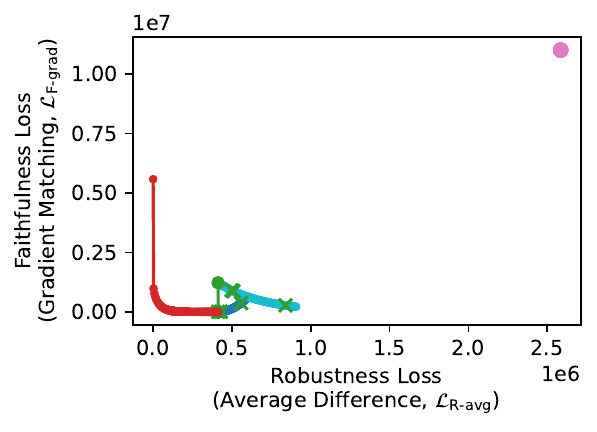}
        \end{minipage}%
        \begin{minipage}{.33\linewidth}
            \centering
            {\tiny\sffamily\textbf{\hspace{3pt}Quasi-Periodic\\[-\baselineskip]\hspace{3pt}with Exponentiated Inputs}} \\[-1pt]
            \includegraphics[width=\linewidth,trim={0 0 0 9pt},clip]{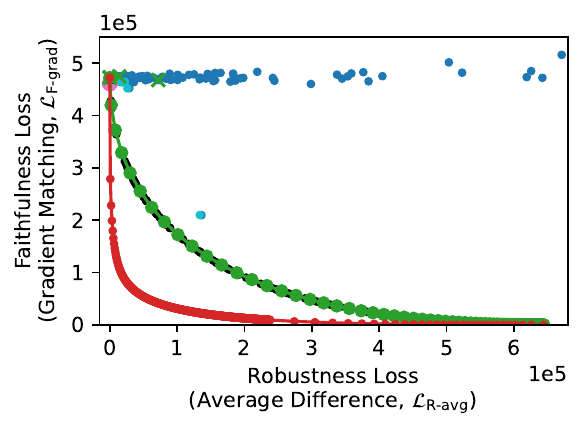}
        \end{minipage}%
        \begin{minipage}{.33\linewidth}
            \centering
            {\tiny\sffamily\textbf{\hspace{3pt}Quasi-Periodic\\[-\baselineskip]\hspace{3pt}with Quadratic Term}} \\[-1pt]
            \includegraphics[width=\linewidth,trim={0 0 0 9pt},clip]{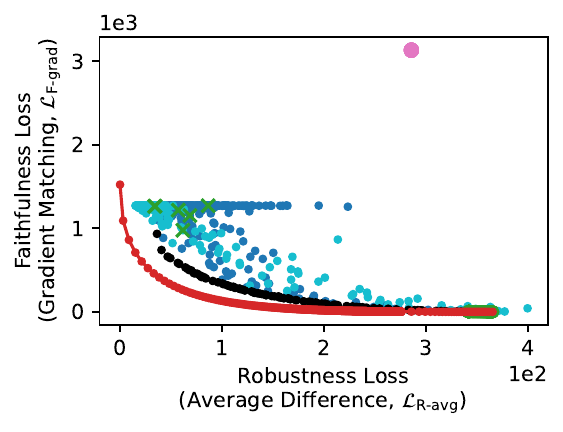}
        \end{minipage}
    \end{minipage}%

    {\tiny\sffamily{} Robustness Loss (Average Difference, $\mathcal{L}_{\texttt{R-avg}}$)}%
    \caption{\textbf{Faithfulness vs.\ Robustness.} ss. Comparison of POE and baselines (SmoothGrad, LIME, AGG, and MOFAE) with SHAP. Due to its lack of tunable hyperparameter, SHAP provides only a single explanation, which is not optimal for all the functions above compared to our methods and other baselines.}
    \label{fig:transductive-duo-shap}
\end{figure}

\subsection{Higher Dimensional Experiments}
\subsubsection{
POLYNOMIALS AND PERIODIC FUNCTIONS}
Below we have additional results for polynomials and periodic functions for $D=10$. We show that
our method consistently provides optimal explanations with an option to control the trade-off between faithfulness
and robustness. For $D=10$ inputs, we randomly sampled 100 points from a uniform distribution \( U(-5, 5) \). 
\begin{figure}[h!]
     \begin{subfigure}[b]{0.28\linewidth}
            \centering
            {\tiny\sffamily\textbf{\hspace{3pt}Quadratic}} \\[-1pt]
            \includegraphics[width=\linewidth]{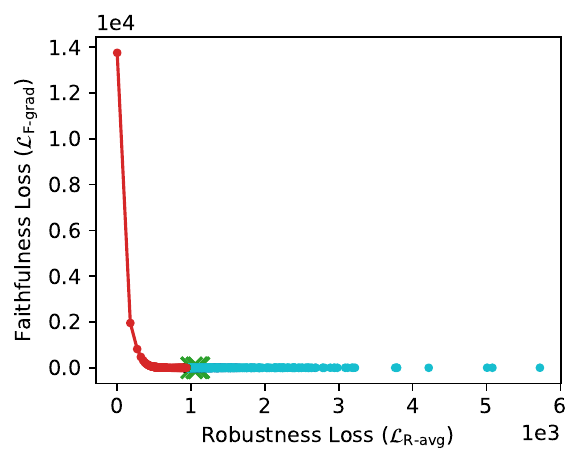}
            \caption{}
            \label{fig:5a}
        \end{subfigure}%
        % \begin{subfigure}[b]{0.35\linewidth}
        %     \centering
        %     \includegraphics[width=\linewidth]{figures/transductive_10D/quadratic_10D.pdf}
        %     \caption{}
        %     \label{fig:5b}
        % \end{subfigure}%
        \begin{subfigure}[b]{0.27\linewidth}
            \centering
            {\tiny\sffamily\textbf{\hspace{3pt}Periodic}} \\[-1pt]

            \includegraphics[width=\linewidth]{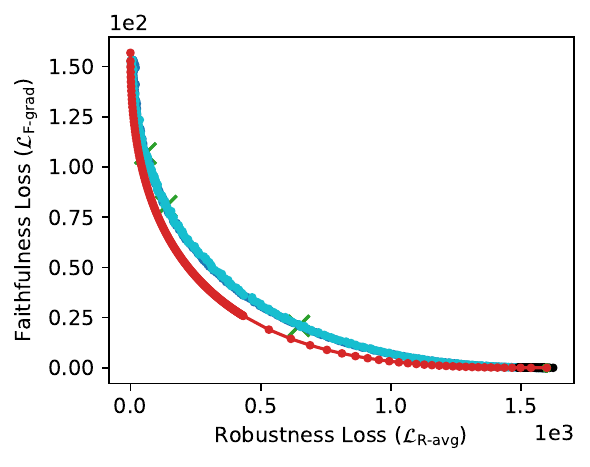}
            \caption{}
            \label{fig:5c}
        \end{subfigure}%
        \begin{subfigure}[b]{0.27\linewidth}
            \centering
            {\tiny\sffamily\textbf{\hspace{3pt}Quasi-Periodic with Quadratic Term}} \\[-1pt]

            \includegraphics[width=\linewidth]{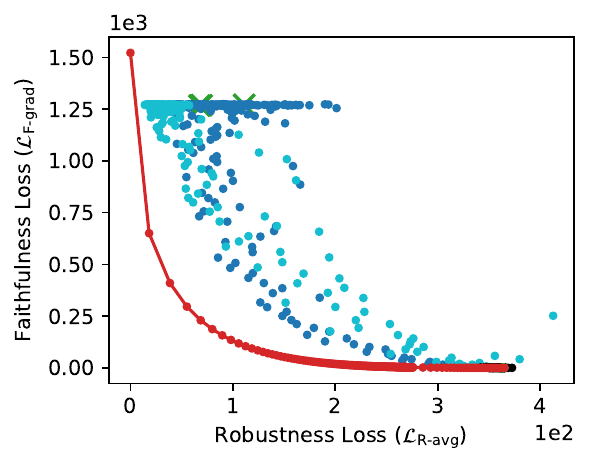}
            \caption{}
            \label{fig:5d}
        \end{subfigure}%
        % \begin{subfigure}[b]{0.35\linewidth}
        %     \centering
        %     \includegraphics[width=\linewidth]{figures/transductive_10D/quasi_2_10D.pdf}
        %     \caption{}
        %     \label{fig:5e}
        % \end{subfigure}%
        \begin{subfigure}[b]{0.27\linewidth}
            \centering
            {\tiny\sffamily\textbf{\hspace{3pt}Quasi-Periodic with Linear Term}} \\[-1pt]
            \includegraphics[width=\linewidth]{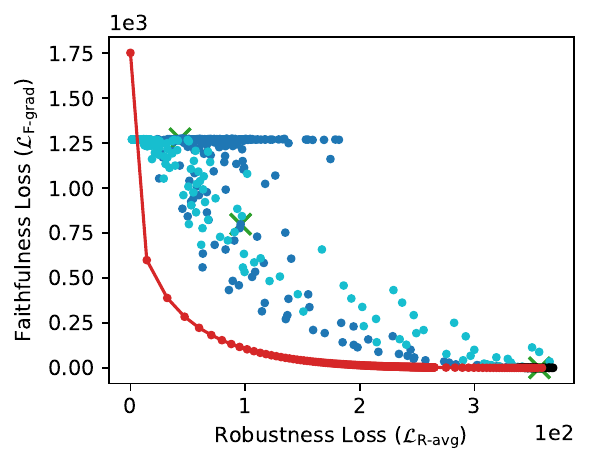}
            \caption{}
            \label{fig:5f}
        \end{subfigure}%

    \caption{\textbf{Faithfulness vs.\ Robustness.}Trade-off between faithfulness loss and robustness loss for various  $\lambda$  values, the hyperparameter controlling trade-off between faithfulness and robustness in our method and AGG.}
    %\textcolor{red}{Hiwot: I will add the zoomed in version of e }
    \label{fig:5}
\end{figure}

\subsubsection{Experiment with Images}

We compared our method to baselines in an experiment using a pretrained ResNet model~\cite{he2016deep}  which contains 25,557,032 parameters on 10 different input images from ImageNet~\cite{russakovsky2015imagenet}

\begin{figure}[H]
    \centering
    \includegraphics[width=0.5\linewidth]{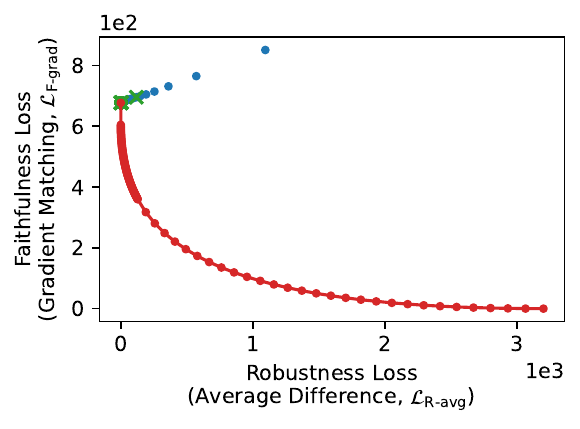}
    \caption{\textbf{Faithfulness vs.\ Robustness.}We observe that our method provides explanations that are Pareto-optimal and capable of managing trade-offs between properties. In contrast, the baselines produce  explanations that are not optimal and concentrated in a limited region of the Pareto front(upper left corner).}
    \label{fig:enter-label}
\end{figure}
\vspace{-10.0 em} 
\subsection{Neural Network}
We trained  neural network models with one hidden layer using the \cite{solar_flare_89} dataset. The results show that our method provides more optimal solutions than the baselines. Furthermore, they highlight the limitations of AGG and MOFAE, as these methods are highly dependent on the quality of the base explanations used to generate the explanations. 

\begin{figure}[H] 
    \centering
    \begin{tabular}{cccc}

        % Row 1
        \includegraphics[width=0.22\textwidth]{{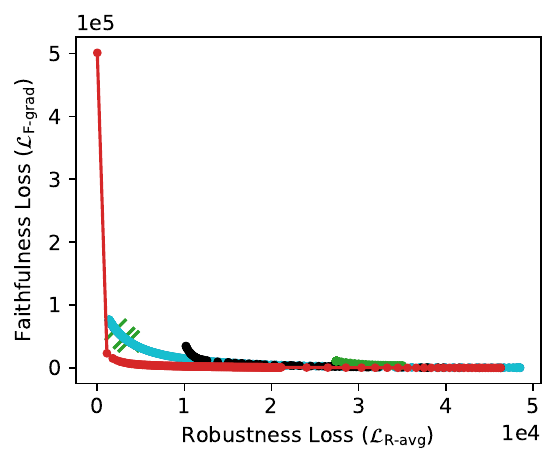}} &
        \includegraphics[width=0.22\textwidth]{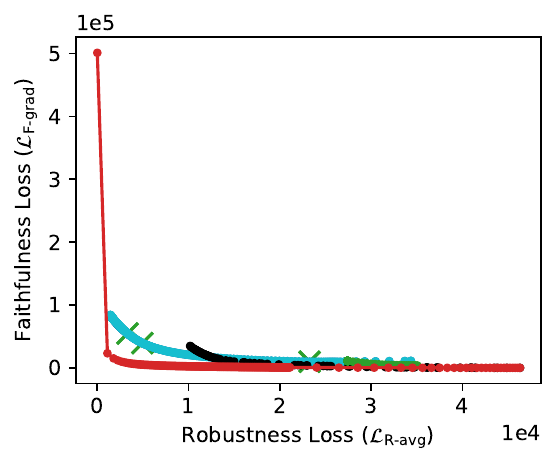} &
        \includegraphics[width=0.22\textwidth]{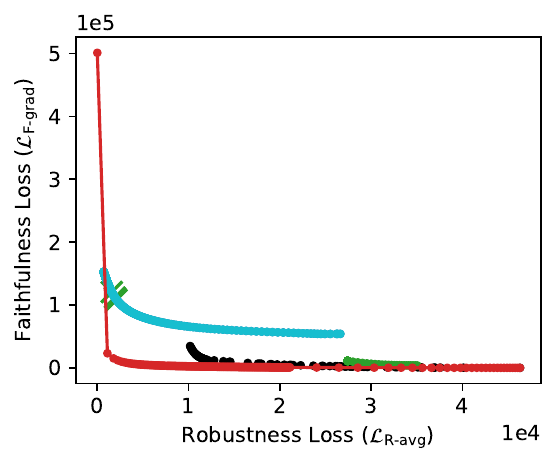} &
        \includegraphics[width=0.22\textwidth]{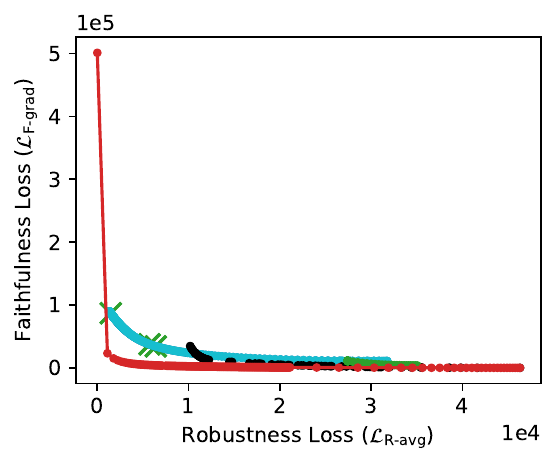} \\

        % Row 2
        \includegraphics[width=0.22\textwidth]{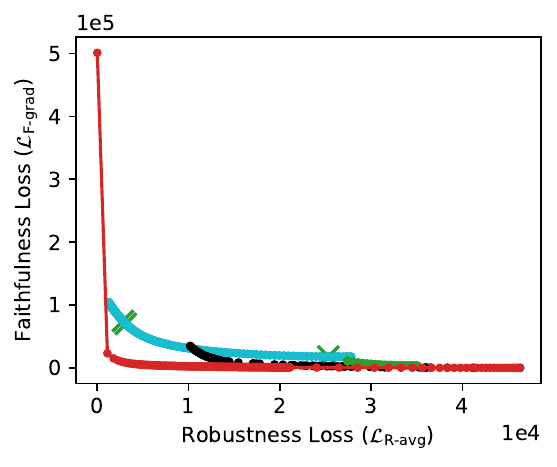} &
        \includegraphics[width=0.22\textwidth]{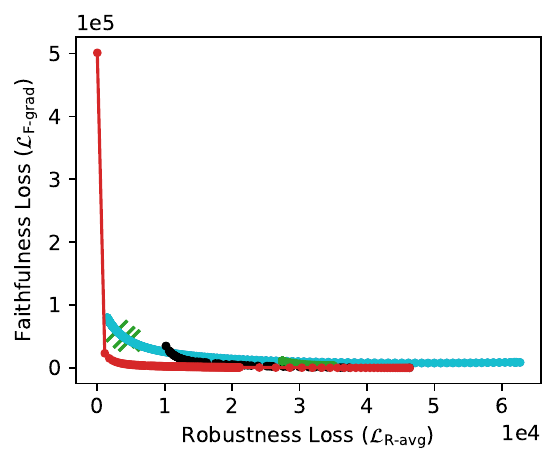} &
        \includegraphics[width=0.22\textwidth]{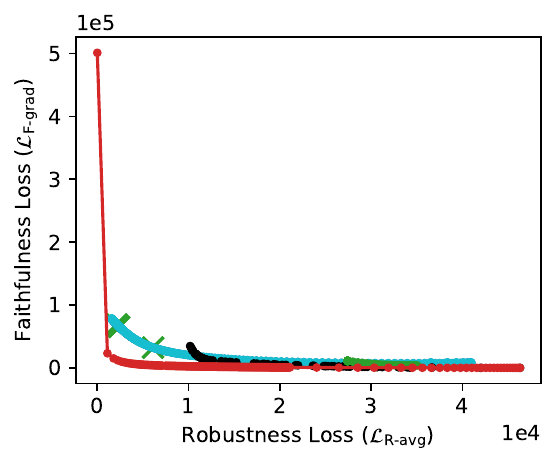} &
        \includegraphics[width=0.22\textwidth]{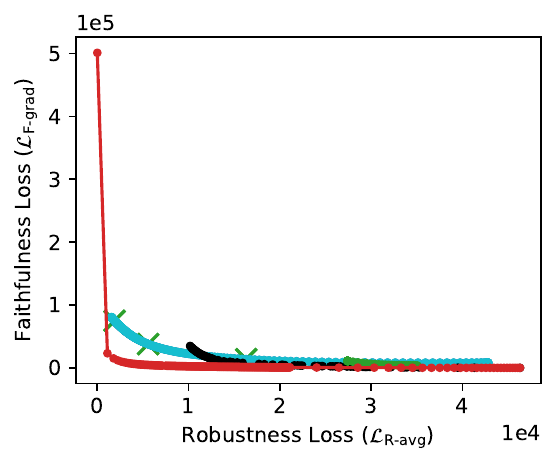} \\

        % Row 3
        \includegraphics[width=0.22\textwidth]{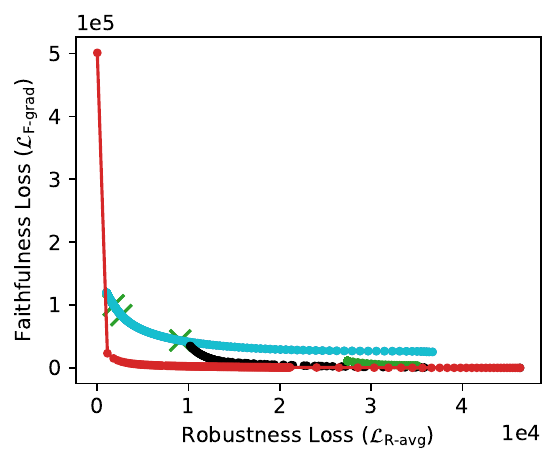} &
        \includegraphics[width=0.22\textwidth]{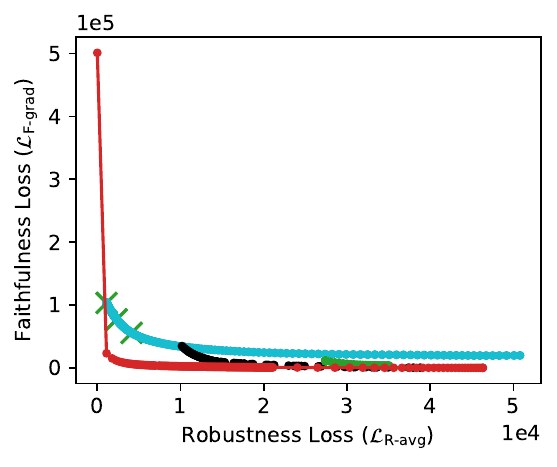} &
        \includegraphics[width=0.22\textwidth]{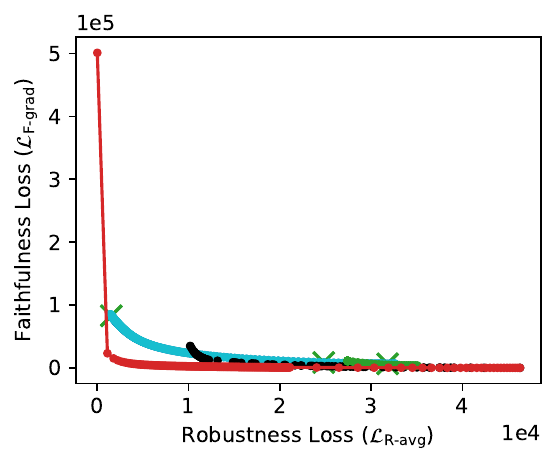} &
        \includegraphics[width=0.22\textwidth]{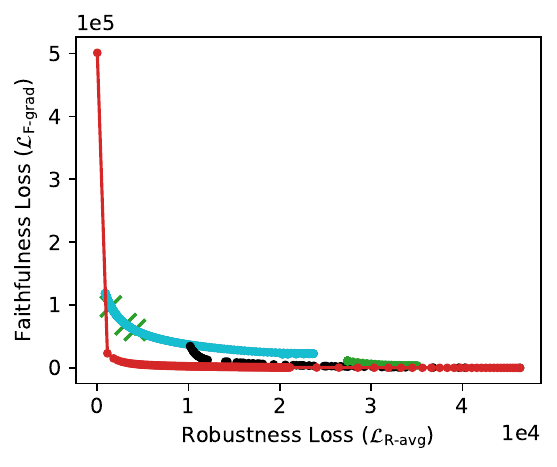} \\
        % Row 4 
        \includegraphics[width=0.22\textwidth]{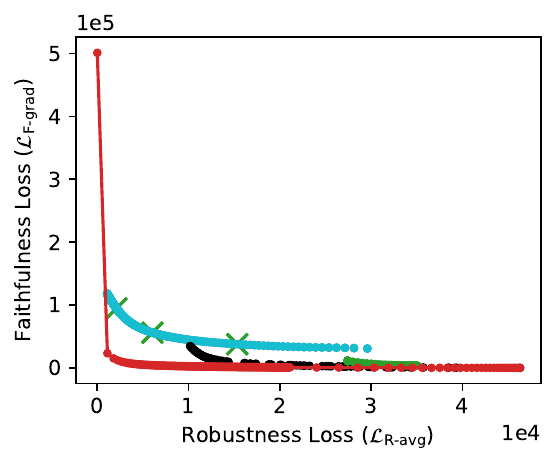} &
        \includegraphics[width=0.22\textwidth]{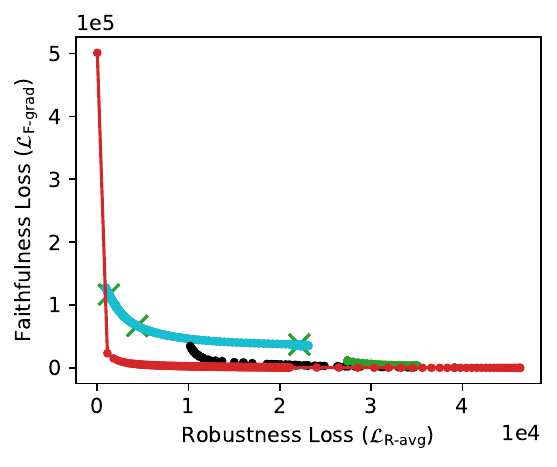} &
        \includegraphics[width=0.22\textwidth]{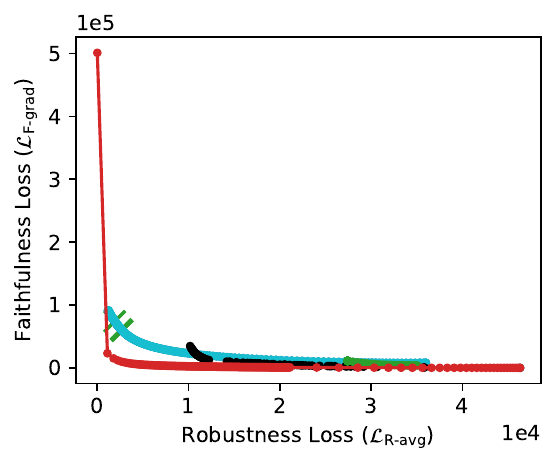} &
        \includegraphics[width=0.22\textwidth]{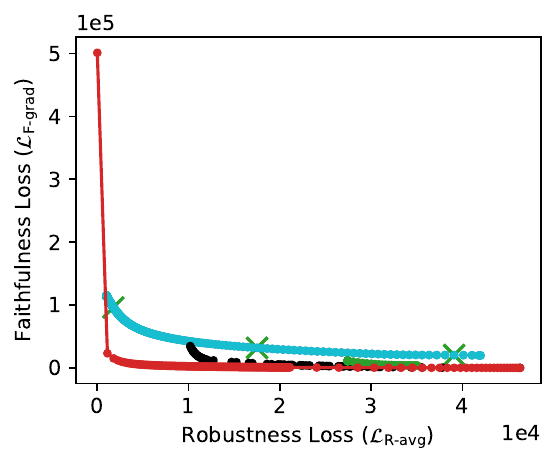} \\
        % Row 5
        \includegraphics[width=0.22\textwidth]{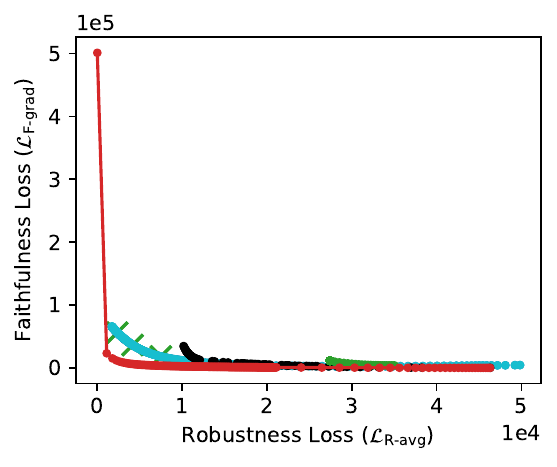} &
        \includegraphics[width=0.22\textwidth]{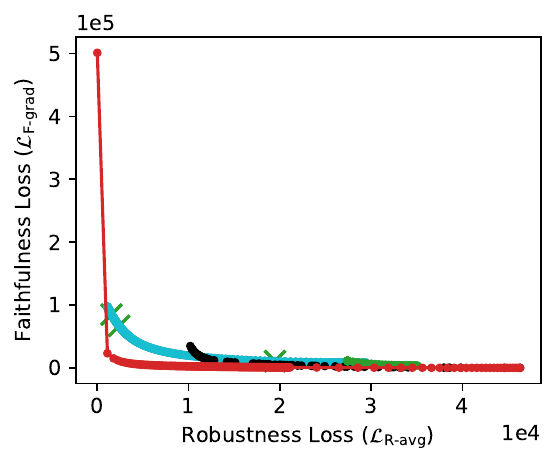} &
        \includegraphics[width=0.22\textwidth]{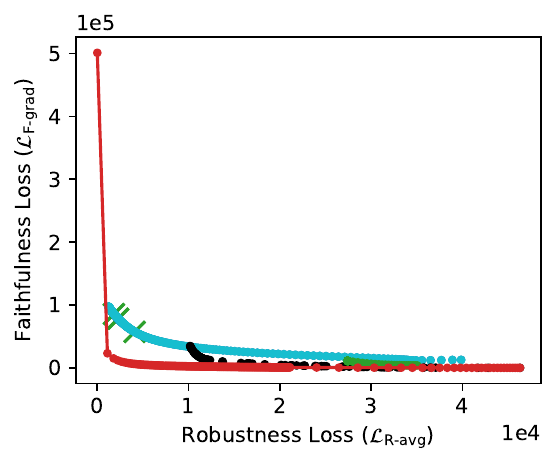} &
        \includegraphics[width=0.22\textwidth]{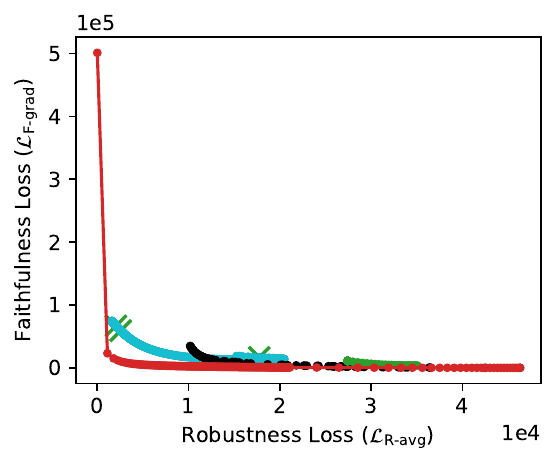} \\
        % Row 6 
        \includegraphics[width=0.22\textwidth]{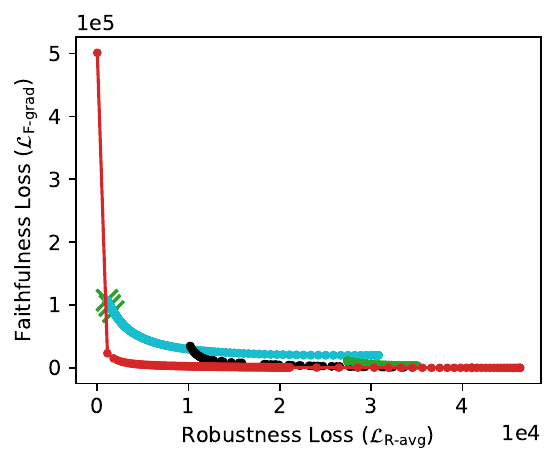} &
        \includegraphics[width=0.22\textwidth]{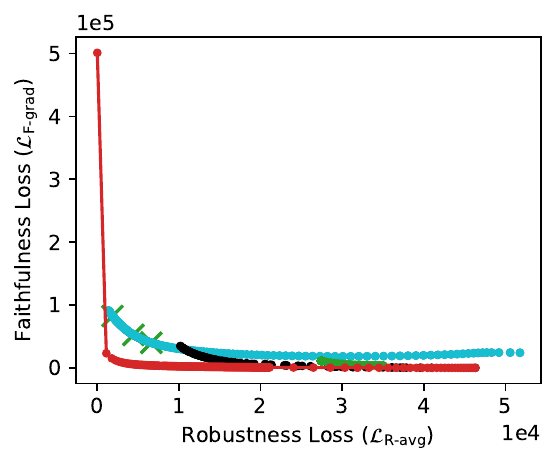} &
        \includegraphics[width=0.22\textwidth]{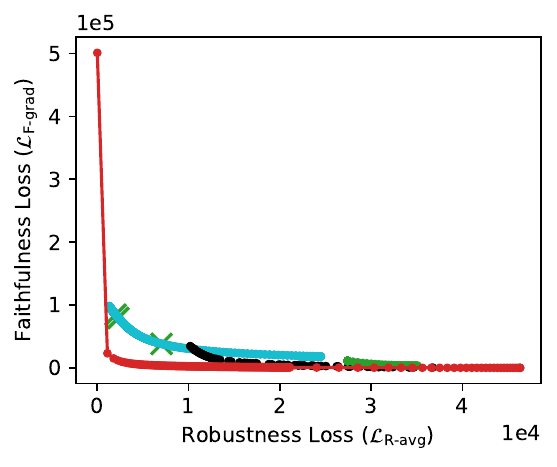} &
        \includegraphics[width=0.22\textwidth]{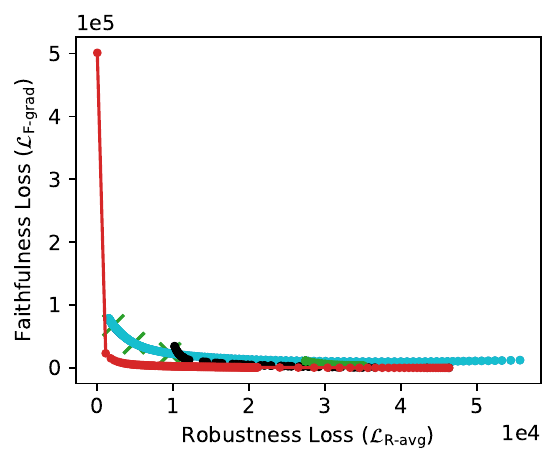} \\
    \end{tabular}
    \label{fig:image_grid_1}
\end{figure}

% \clearpage

\begin{figure}[H] 
    \centering
    \begin{tabular}{cccc}

        % Row 1
        \includegraphics[width=0.22\textwidth]{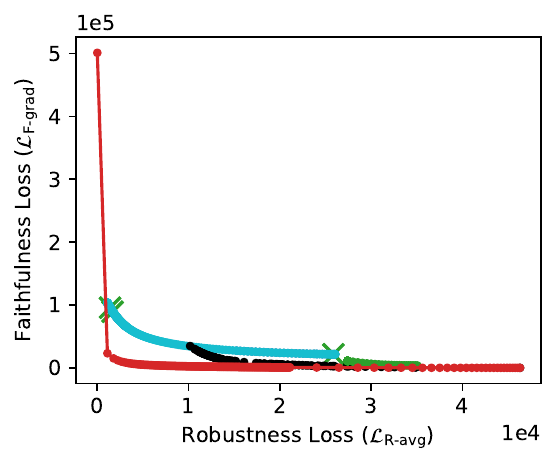} &
        \includegraphics[width=0.22\textwidth]{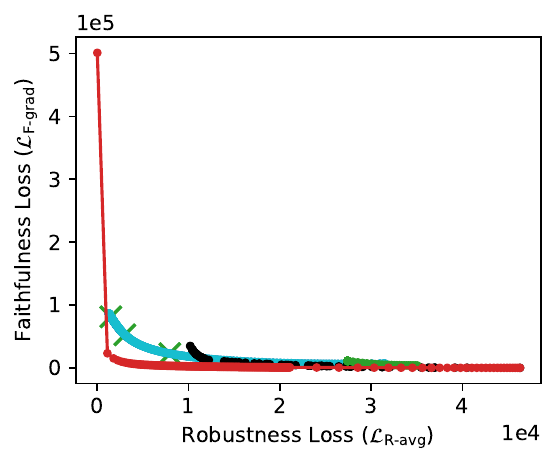} &
        \includegraphics[width=0.22\textwidth]{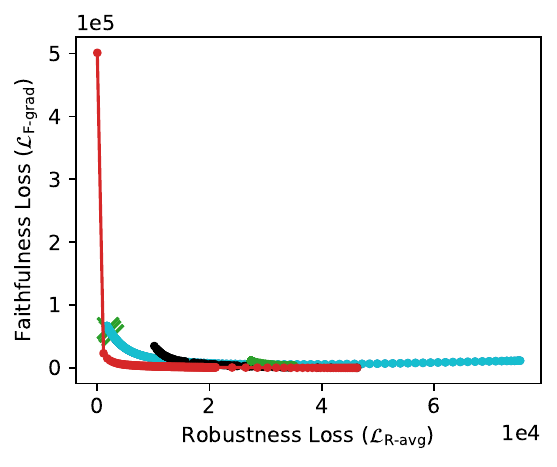} &
        \includegraphics[width=0.22\textwidth]{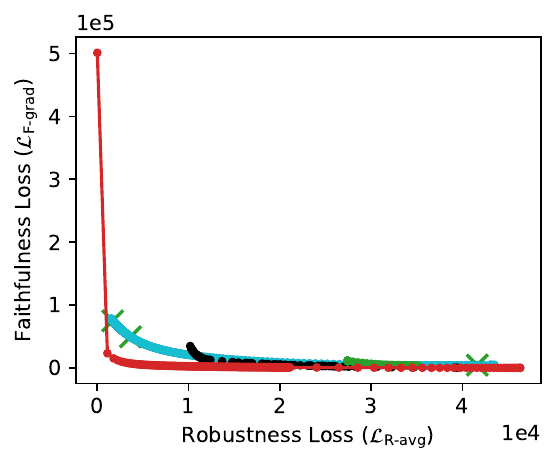} \\

        % Row 2
        \includegraphics[width=0.22\textwidth]{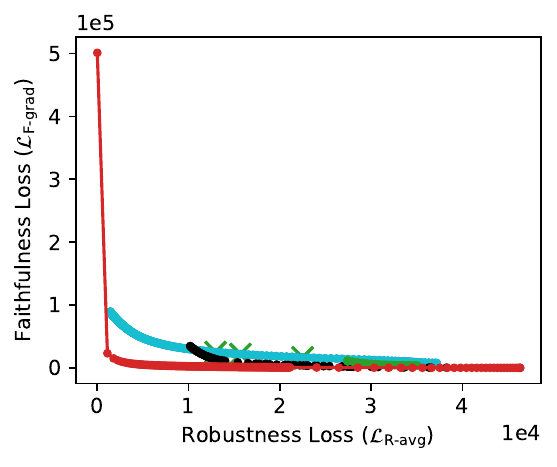} &
        \includegraphics[width=0.22\textwidth]{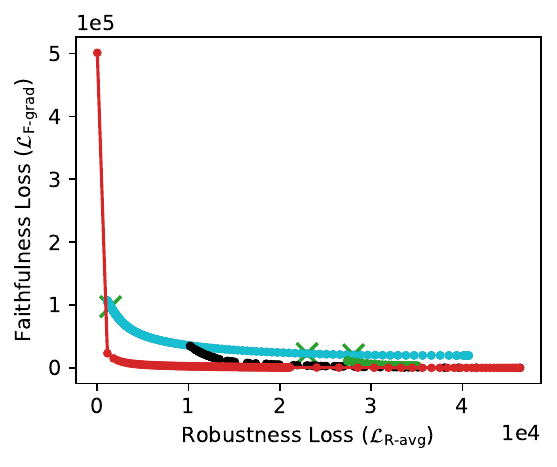} &
        \includegraphics[width=0.22\textwidth]{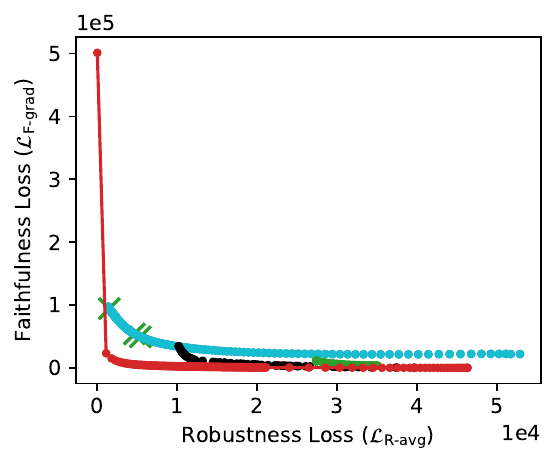} &
        \includegraphics[width=0.22\textwidth]{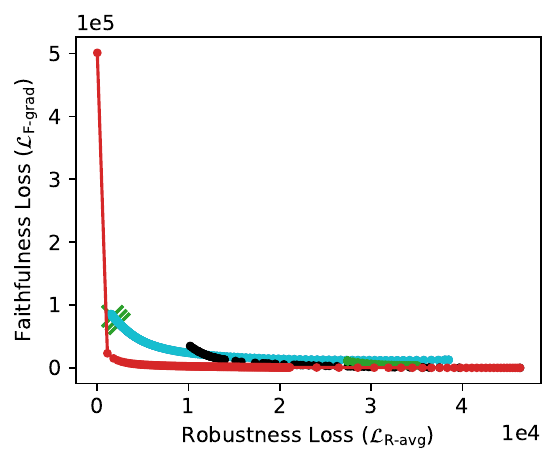} \\

        % Row 3
        \includegraphics[width=0.22\textwidth]{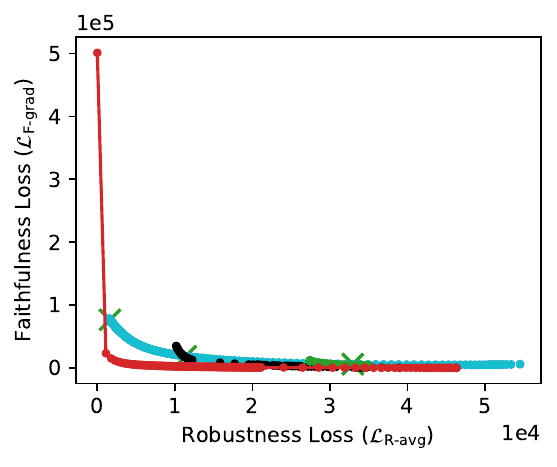} &
        \includegraphics[width=0.22\textwidth]{figures-scipy/neural_network_33.pdf} &
        \includegraphics[width=0.22\textwidth]{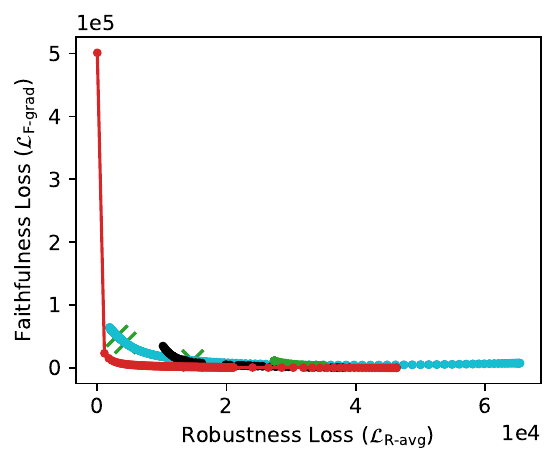} &
        \includegraphics[width=0.22\textwidth]{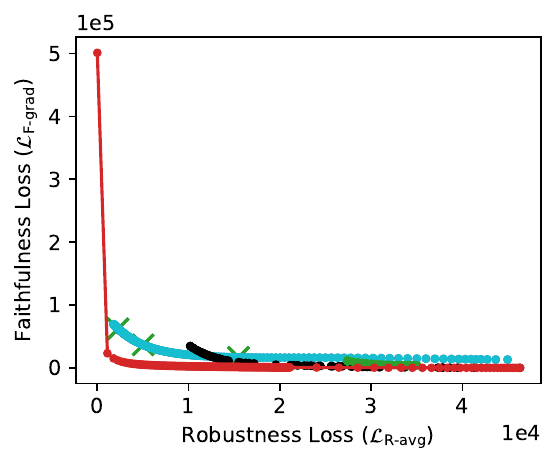} \\
        % Row 4 
        \includegraphics[width=0.22\textwidth]{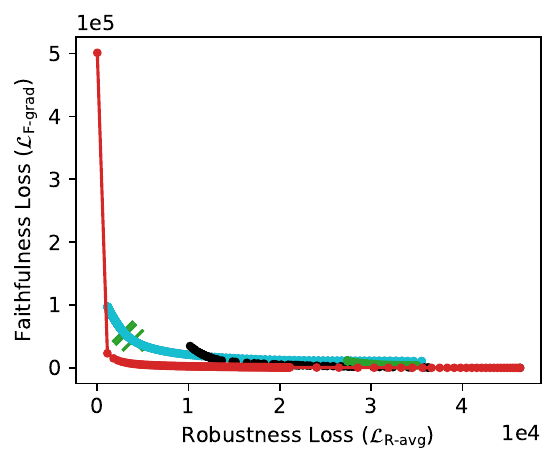} &
        \includegraphics[width=0.22\textwidth]{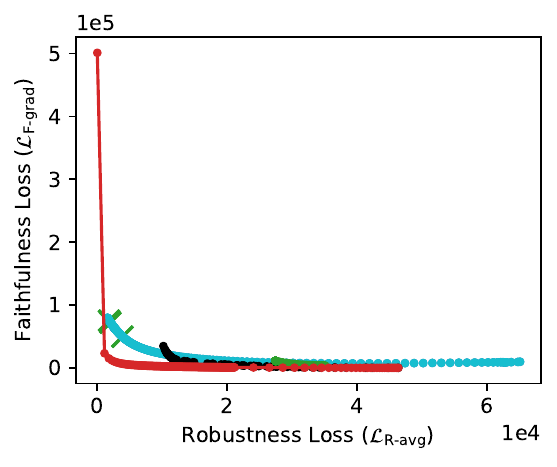} &
        \includegraphics[width=0.22\textwidth]{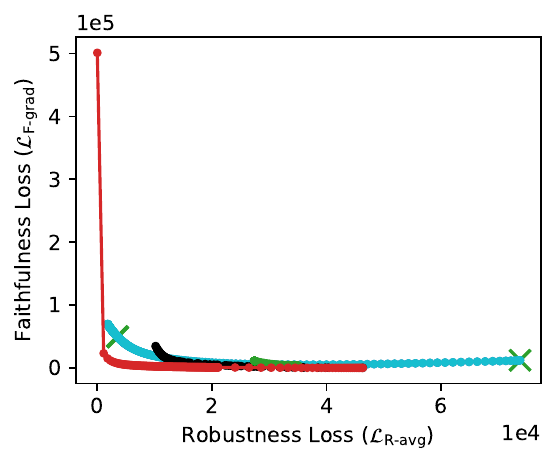} &
        \includegraphics[width=0.22\textwidth]{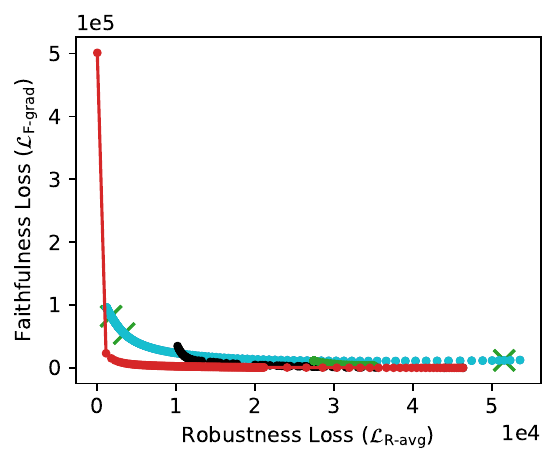} \\
        % Row 5
        \includegraphics[width=0.22\textwidth]{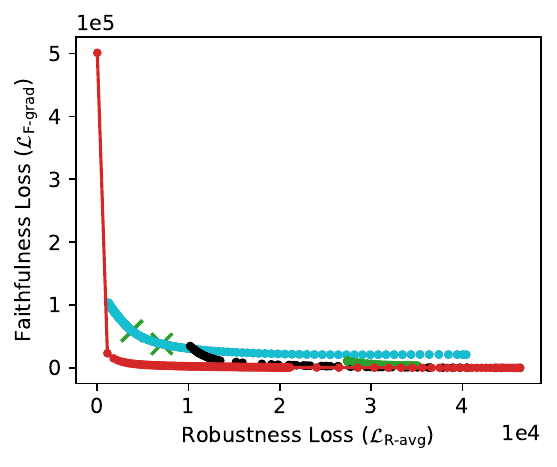} &
        \includegraphics[width=0.22\textwidth]{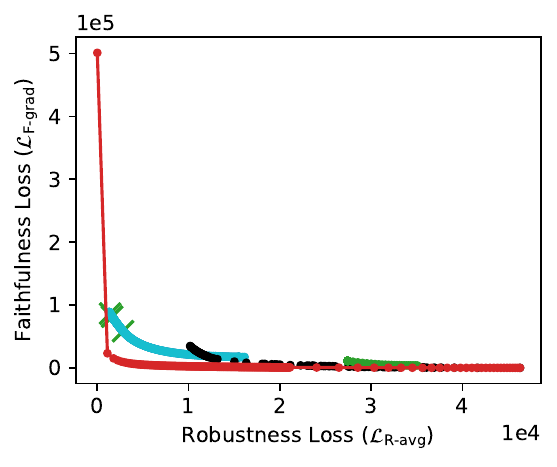} &
        \includegraphics[width=0.22\textwidth]{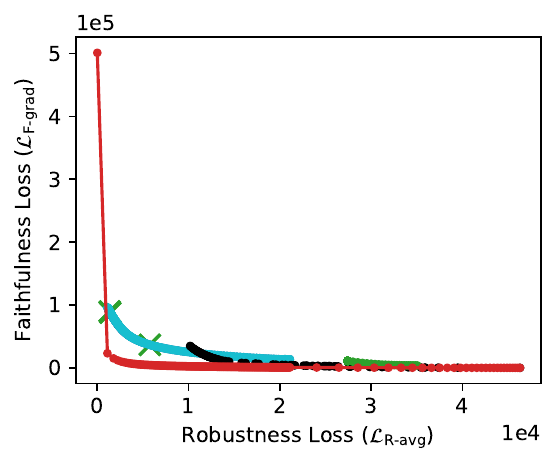} &
        \includegraphics[width=0.22\textwidth]{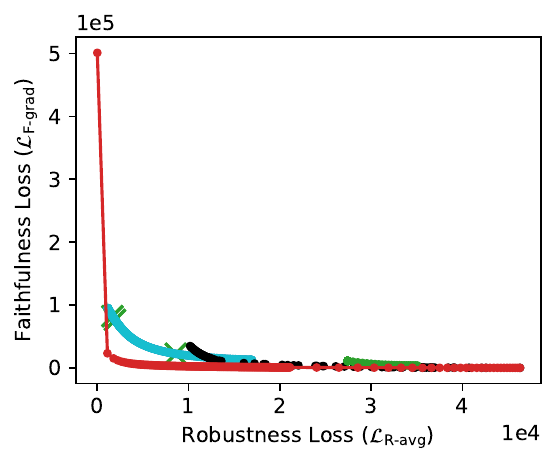} \\
        % Row 6 
        \includegraphics[width=0.22\textwidth]{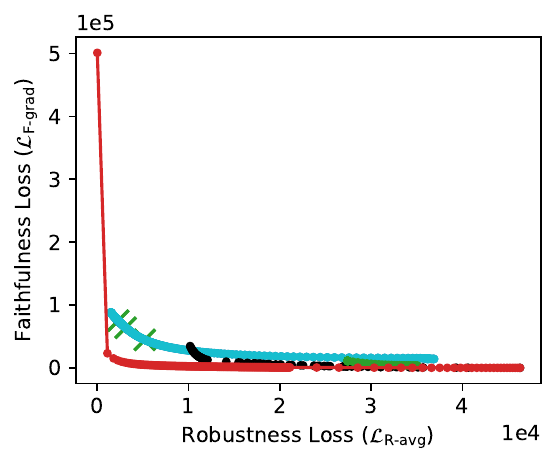} &
        \includegraphics[width=0.22\textwidth]{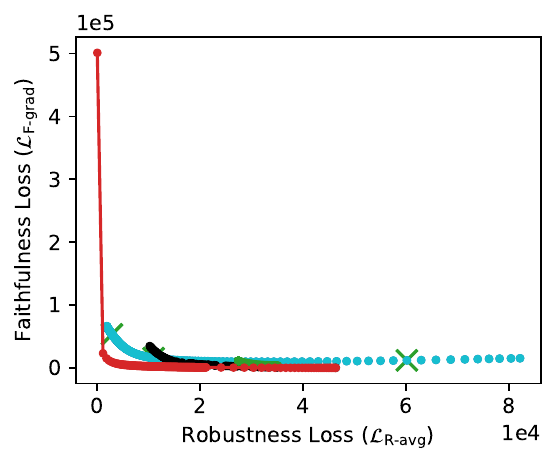} &
        \includegraphics[width=0.22\textwidth]{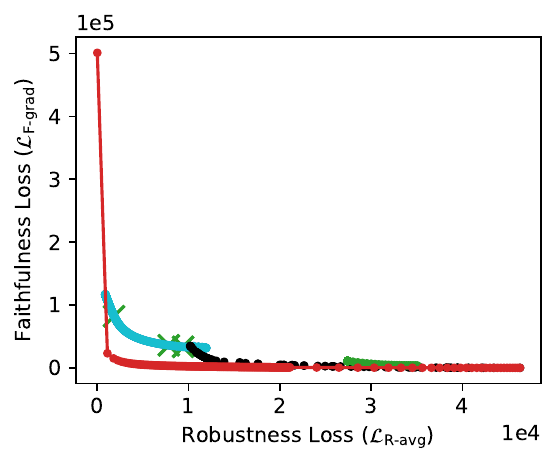} &
        \includegraphics[width=0.22\textwidth]{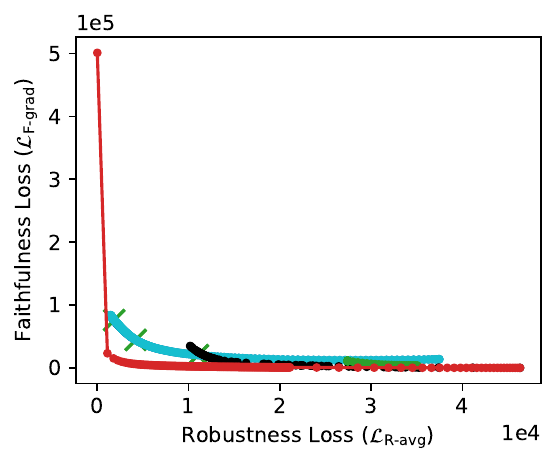} \\
    \end{tabular}
    \label{fig:image_grid_1}
\end{figure}

\section{Qualitative Comparison }
\begin{figure}[H]
    \centering
    \includegraphics[width=1.0\linewidth]{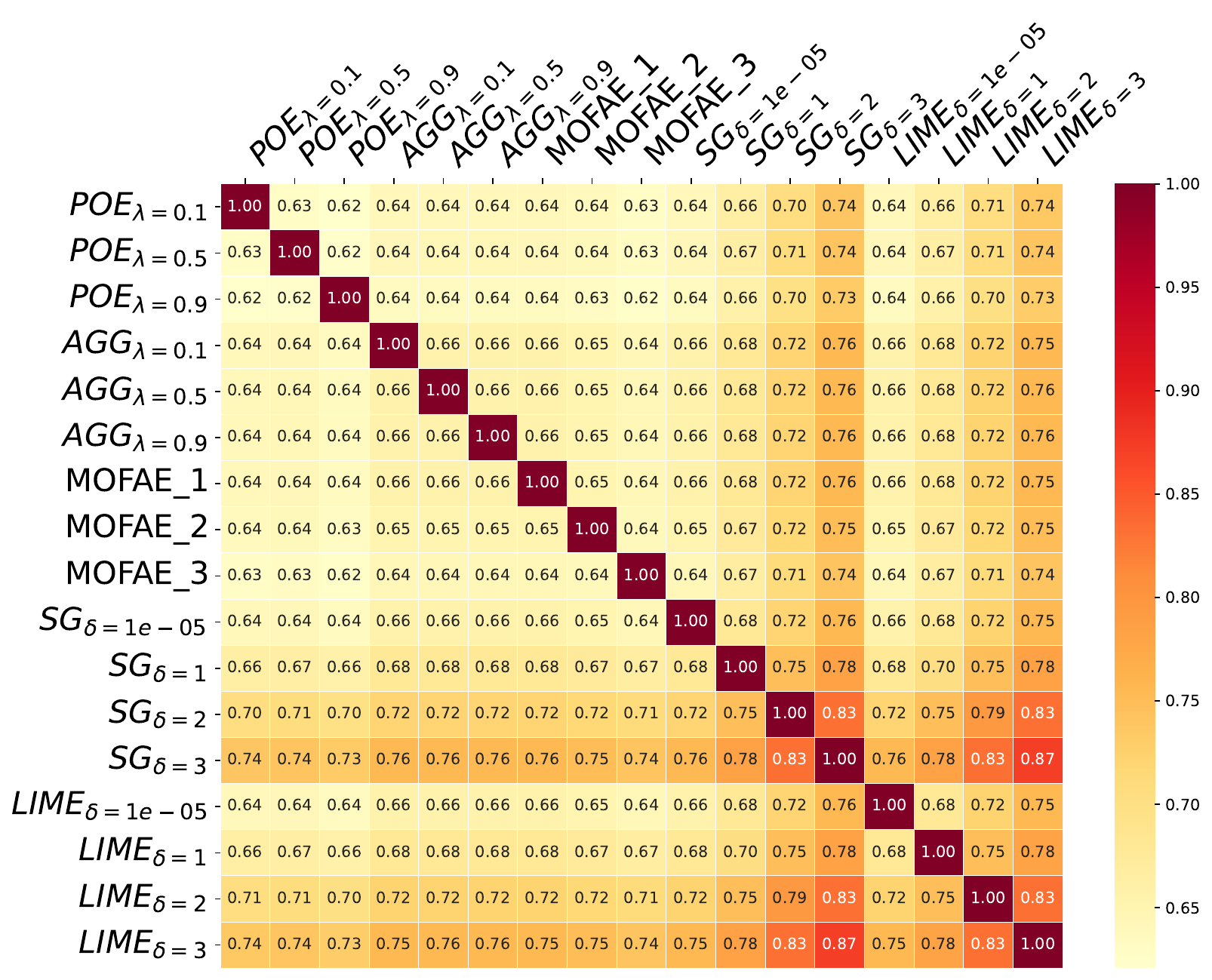}
    \caption{Comparison of POE with baselines with respective hyperparameters using cosine similarity for a cubic function}
    \label{fig:cosine_simialrity}
\end{figure}

\begin{figure}[H]
    \centering
    \includegraphics[width=1.0\linewidth]{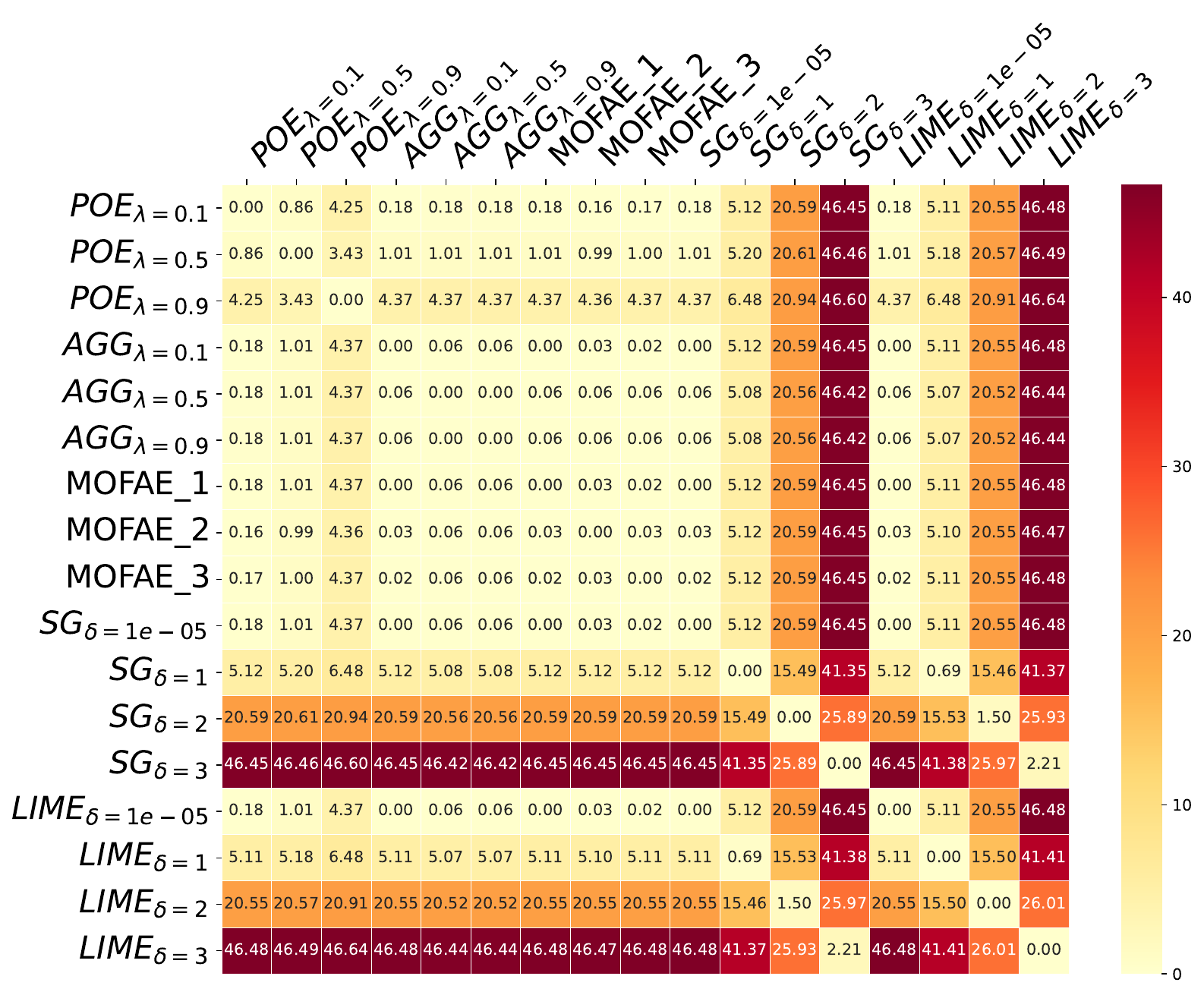}
    \caption{Comparison of POE with baselines with respective hyperparameters using $L_2$ distance for a cubic function.}
    \label{fig:l2-distance}
\end{figure}

We compared POE with baselines based on agreement on the top-1 most important feature, where a score closer to 1 indicates stronger agreement. Our method demonstrates consistently high agreement across varying trade-off hyperparameter values ($\lambda$), whereas the baselines show comparably lower agreement scores regardless of changes to their respective hyperparameters. In addition, we also compared POE with baselines based on cosine similarity and $L_2$ distance, results are shown in Figure~\ref{fig:cosine_simialrity} and \ref{fig:l2-distance}. 

\begin{figure}[H]
    \centering
    \includegraphics[width=1.0\linewidth]{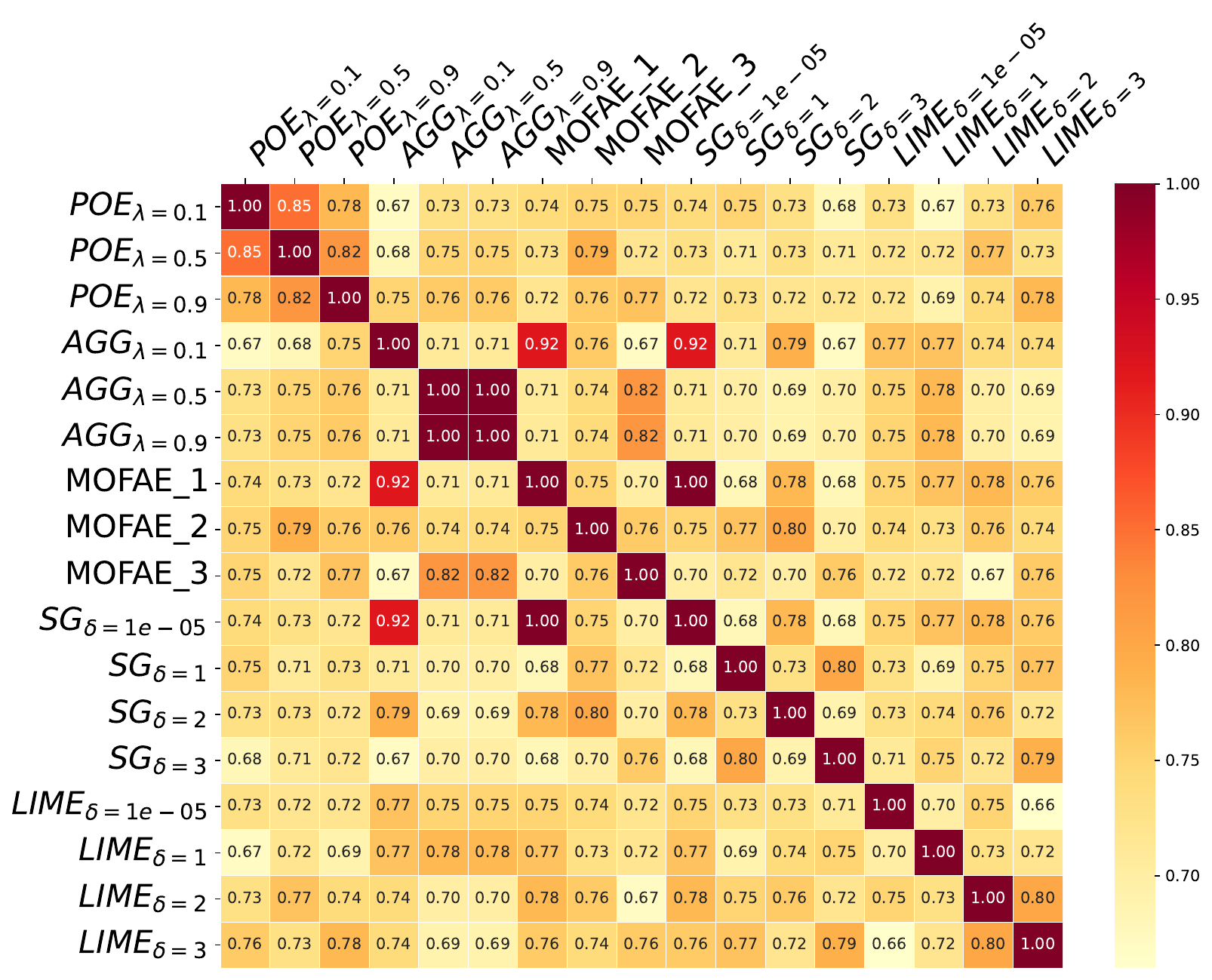}
    \caption{Comparison of POE with baselines using Agreement on the top feature for a cubic function.}
    \label{fig:top-feature-comparision}
\end{figure}

\section {Scalability of Transductive Setting}
In the transductive setting, our method is faster, providing an explanation in approximately 0.00 seconds per lambda for lower-dimensional inputs. In comparison, the baselines require substantially more time: SmoothGrad (~0.39 seconds), LIME (~0.13 seconds), AGG (~33.01 seconds), and MOFAE (~40.51 seconds) per respective hyperparameter. This difference becomes more pronounced with higher-dimensional inputs, such as images.

The scalability of our method, as well as that of some baseline methods like AGG and MOFAE, remains a limitation in this version of the work. Our primary goal was to demonstrate the generality and usefulness of the idea of treating explanations as quantities. Developing more efficient algorithms for scalability, including amortization schemes, is an important direction for future research.

\section{Further Qualitative Evaluation}
\begin{figure}[h]
    \centering
    \includegraphics[width=0.66\linewidth]{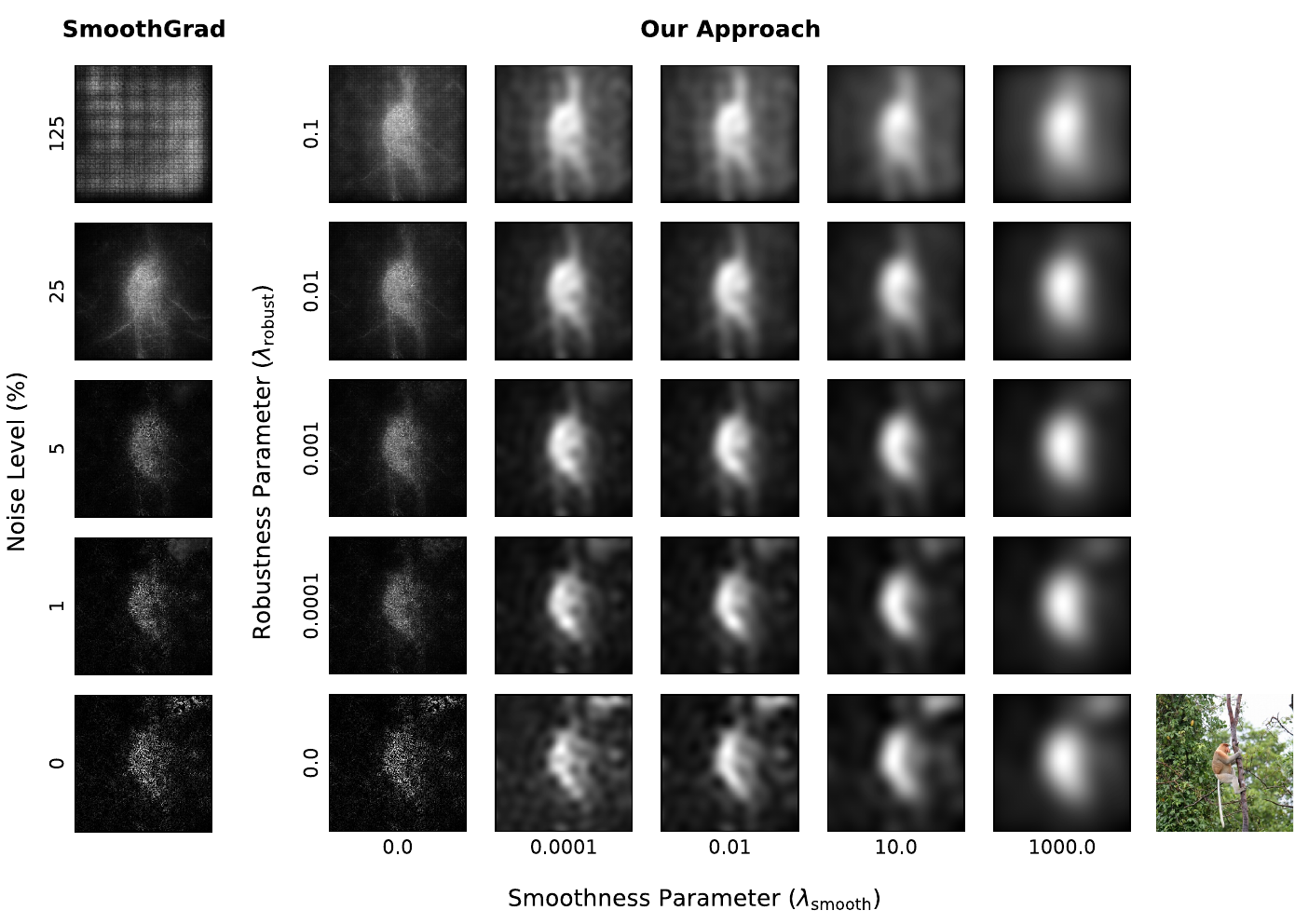}
    \caption{Comparison of our approach vs.\ SmoothGrad for a different input image when explaining image-based models.}
    \label{fig:qualitative-example1}
\end{figure}

In Figure~\ref{fig:qualitative-example0}, we argued that our approach can control the balance between faithfulness, robustness and smoothness while highlighting that SmoothGrad can only manage the trade-off between faithfulness and robustness (missing smoothness). Here, we present a similar figure but for a different initial image (Figure~\ref{fig:qualitative-example1}) to show that this pattern is not limited to a single instance.

\section{CHOICE OF INDUCING POINTS}

Our inductive approach required specifying a set of inducing points to be used as training data in GP inference. In Figure~\ref{fig:inductive}, we investigated how the number of these inducing points affect the results of our method and concluded that solutions approach to their transductive counterparts exponentially fast as the number of points increase. During those experiments, we sampled points uniformly at random from the target function's input domain $\Omega$ as our inducing points and used the same set of inducing points for each query point. However, this leaves the question of whether one can be more efficient with their choice of inducing points than uniform sampling. For instance, given a particular query point $\bm{x}^*$, would it be as effective to sample fewer points around $\bm{x}^*$ as oppose to covering the entire input domain $\Omega$?

\looseness-1
\textbf{Setup.}
We consider the same functions%
\footnote{That is with the exception of \textit{quasi-periodic with exponent} as it oscillates more frequently than the period between inducing points considered in this set of experiments.}
as in Figure~\ref{fig:inductive} but with $D=1$ so that we can compute solutions using a very large density of inducing points over the whole input domain ($N=1000$, uniformly spread over interval $\Omega=[-10,10]$). Using these inducing points, we generate explanations for $100$~query points, uniformly spread over interval $[-5,5]$ (a narrower range to avoid any edge artifacts). These explanations optimize for faithfulness and robustness induced by a Gaussian kernel with length scale $\Lambda=0.25)$, where $\lambda_{\text{Faithful}}=1$ and $\lambda_{\text{Robustness}}=0.01$. Treating these explanations as ground-truth, we consider three other explanations generated using different inducing points: (i) \textit{Global} uses a smaller number of points, $N=100$, but still uniformly spread over the whole input domain; (ii) \textit{Local} also uses $N=100$ points but spread over a narrower interval $\Omega'=[\bm{x}^*-R,\bm{x}^*+R]$ around each query point $\bm{x}^*$; and (iii) \textit{Global+Local} uses the union of both point sets.

Our GP-based approximation to the inductive problem, stated in Proposition~\ref{prop}, requires inducing points $\{\bm{x}_n\}$ to be distributed uniformly over $\Omega$ (as in \textit{Global}). We use importance re-weighting to account for the distribution shift from \textit{Global} to \textit{Local} or \textit{Global+Local}. When inducing points are distributed according to an arbitrary density $p(\bm{x}_n)$, each sample $\bm{x}_n$ is assigned the importance weight $(1/|\Omega)/p(\bm{x}_n)$. For instance, if inducing points are sampled uniformly over $\Omega'\subset\Omega$ instead of $\Omega$ (as in \textit{Local}), then each inducing point is now weighted more heavily with weighting factor $|\Omega'|/|\Omega|>1$.

\begin{figure}[t]
    \centering%
    \includegraphics[width=\linewidth]{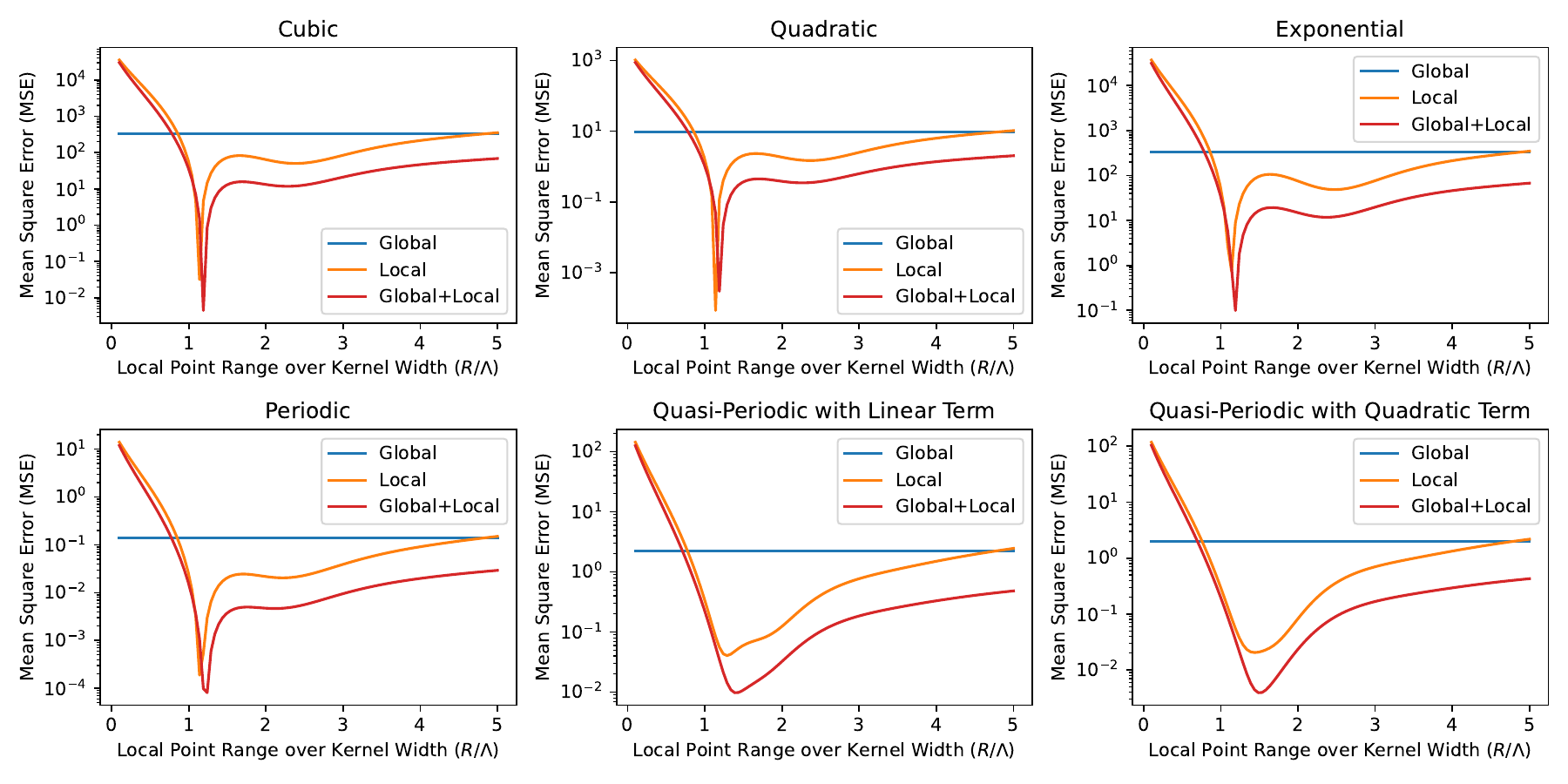}%
    \caption{Comparison of \textit{Global}, \textit{Local}, and \textit{Global+Local} inducing points. As long as the range of local points are wider than the kernel width, they achieve lower error than \textit{Global} using the same number of inducing points. This efficiency is reduced as the local point range gets wider and wider.}%
    \label{fig:global-local}%
\end{figure}

\textbf{Results.}
In Figure~\ref{fig:global-local}, we report the mean square error of $\textit{Global}$, $\textit{Local}$, and $\textit{Global+Local}$ as the range of local inducing points $R$ varies relative to the kernel length scale $\Lambda$. We see that $R\ll \Lambda$ leads to worse accuracy than \textit{Global}, inducing points fail to cover parts of the input domain that are similar to the query point as dictated by our kernel. As long as $R\gtrapprox \Lambda$, \textit{Local} achieves smaller error than \textit{Global} despite using the same number of inducing points. This is because a majority of the inducing points used by \textit{Global} would have a low similarity to a given query point; \textit{Local} avoids this by using inducing points in the immediate vicinity of each query point. However, this efficiency diminishes as $R$ keeps increasing and the inducing points become more spread out.

While \textit{Local} makes more efficient use of inducing points, the fact that the set of inducing points is different for each query point requires performing GP inference from scratch for each individual query point (unlike \textit{Global} where the GP posterior can be computed once for the fixed set of inducing points). Therefore, we give the following practical advice: If query points already cover the input domain densely, \textit{Global} is likely to be the computationally more efficient strategy (despite requiring more computation upfront). However, if query points are sparse relative to the size of the input domain (for instance, if the input domain is very high dimensional), than $\textit{Local}$ with a range $R$ on the same magnitude as the kernel length scale $\Lambda$ is likely to be the more efficient strategy instead.

\end{document}